\RequirePackage{fix-cm}
\documentclass{svjour3}                     % onecolumn (standard format)
\pdfoutput=1
\smartqed  % flush right qed marks, e.g. at end of proof
\usepackage{graphicx}
\usepackage{multicol}
\usepackage{subfigure}
\usepackage{amsmath}
\usepackage{amssymb}
\usepackage{amsfonts}
\usepackage{booktabs} 
\usepackage{enumerate}
\usepackage{xcolor}
\usepackage{color} 
\usepackage{stfloats}
\usepackage{lineno,hyperref}
\usepackage{stmaryrd}
\usepackage{float}
\usepackage{stfloats}
\usepackage{epstopdf}

\usepackage{multirow}
\modulolinenumbers[5]
\usepackage{url}
\usepackage{bm}
\usepackage[ruled,vlined]{algorithm2e}
\graphicspath{{figures/}}

\begin{document}

\title{Edge-Based Blur Kernel Estimation Using Sparse Representation and Self-Similarity \thanks{National Natural Science Foundation of China (61501008) and Beijing Municipal Natural Science Foundation (4172002).}
}
%\subtitle{}

\titlerunning{Edge-Based Blur Kernel Estimation}        % if too long for running head

\author{Jing Yu \and 
Zhenchun Chang \and Chuangbai Xiao
}

\authorrunning{J. Yu} % if too long for running head

\institute{
           J. Yu \and C. Xiao \at 
              Faculty of Information Technology, Beijing University of Technology, Beijing 100124, China \\
              \email{jing.yu@bjut.edu.cn; cbxiao@bjut.edu.cn}           %  \\
           \and
           Z. Chang \at
              Department of Electronic Engineering, Tsinghua University, Beijing 100084, China
}

\date{Received: date / Accepted: date}

\maketitle

\begin{abstract}
Blind image deconvolution is the problem of recovering the latent image from the only observed blurry image when the blur kernel is unknown. In this paper, we propose an edge-based blur kernel estimation method for blind motion deconvolution. In our previous work, we incorporate both sparse representation and self-similarity of image patches as priors into our blind deconvolution model to regularize the recovery of the latent image. Since almost any natural image has properties of sparsity and multi-scale self-similarity, we construct a sparsity regularizer and a cross-scale non-local regularizer based on our patch priors. It has been observed that our regularizers often favor sharp images over blurry ones only for image patches of the salient edges and thus we define an edge mask to locate salient edges that we want to apply our regularizers. Experimental results on both simulated and real blurry images demonstrate that our method outperforms existing state-of-the-art blind deblurring methods even for handling of very large blurs, thanks to the use of the edge mask. 
%\PACS{PACS code1 \and PACS code2 \and more}
\CRclass{10010147 \and 10010371 \and 10010382 \and 10010383}
\end{abstract}

\section{Introduction}

Motion blur caused by camera shake has been one of the most common artifacts in digital imaging. Blind image deconvolution is an inverse process that attempts to recover the latent (unblurred) image from the observed blurry image when the blur kernel is unknown. In general, for most of the work, the degradation is assumed that the observed image is the output of a linear shift invariant (LSI) system to which noise is added.
% You must have at least 2 lines Singhin the paragraph with the drop letter

If the blur is shift-invariant, it can be modeled as the 2-D convolution of the latent image with the blur kernel:
\begin{equation}
    \boldsymbol{y}=\boldsymbol{h}*\boldsymbol{x}+\boldsymbol{v},
    \label{eq:BlurModel}
\end{equation}
where $*$ stands for the 2-D convolution operator, $\boldsymbol{y}$ is the observed blurry image, $\boldsymbol{h}$ is the blur kernel (or point spread function), $\boldsymbol{x}$ is the latent image and $\boldsymbol{v}$ is the additive noise. Then, removing the blur from the observed blurry image becomes a deconvolution operation. When the blur kernel is unknown, the blind deconvolution is a more severely ill-posed inverse problem. The key to the solution of the ill-posed inverse problem is proper incorporation of various image priors about the latent image into the blind deconvolution process. Non-blind image deconvolution seeks an estimate of the latent image assuming the blur is known. In contrast, blind image deconvolution tackles the more difficult, but realistic, problem where the degradation is unknown.

Despite over three decades of research in the field, blind deconvolution still remains a challege for real-world photos with unknown kernels. Recently, blind deconvolution has received renewed attention since Fergus et al.'s work \cite{FergusSingh} and impressive progress has been made in removing motion blur only given a single blurry image. Some methods explicitly or implicitly exploit edges for kernel estimation \cite{Jia,JoshiSzeliski,ChoLee,XuJia}. This idea was introduced by Jia \cite{Jia}, who used an alpha matte to estimate the transparency of blurred object boundaries and performed the kernel estimation using transparency. Joshi et al. \cite{JoshiSzeliski} predict sharp edges using edge profiles and estimate the blur kernel from the predicted edges. However, their goal is to remove small blurs, for it is not trivial to directly restore sharp edges from a severely blurred image. In \cite{ChoLee,XuJia}, strong edges are predicted from the latent image estimate using a shock filter and gradient thresholding, and then used for kernel estimation. Unfortunately, the shock filter could over-sharpen image edges, and is sensitive to noise, leading to an unstable estimate.

Another family of methods exploit various sparse priors for either the latent image $\boldsymbol{x}$ or the motion blur kernel $\boldsymbol{h}$, and formulate the blind deconvolution as a joint optimization problem with some regularizations on both $\boldsymbol{x}$ and $\boldsymbol{h}$ \cite{FergusSingh,LevinWeiss2009,LevinWeiss2011,ShanJia,PerroneFavaro,PerroneFavaro2016}:
\begin{equation}
        (\hat{\boldsymbol{x}},\hat{\boldsymbol{h}}) = \arg\min\limits_{\boldsymbol{x},\boldsymbol{h}}\Big\{\sum\limits_{*}\omega_{*}\Vert\partial_{*}\boldsymbol{y}-\boldsymbol{h}*\partial_{*}\boldsymbol{x}\Vert_2^2  + \lambda_x\rho(\boldsymbol{x}) + \lambda_h\rho(\boldsymbol{h})\Big\},
\end{equation}
where $\partial_{*}\in\{\partial_{0},\partial_{x},\partial_{y},\partial_{xx},\partial_{xy},\partial_{yy},\cdots\}$ denotes the partial derivative operator in different directions and orders, $\omega_{*}$ is a weight for each partial derivative, $\rho(\boldsymbol{x})$ is a regularizer on the latent sharp image $\boldsymbol{x}$, $\rho(\boldsymbol{h})$ is a regularizer on the blur kernel $\boldsymbol{h}$, and $\lambda_x$ and $\lambda_h$ are regularization weights. The first term in the energy minimization formulation of blind deconvolution uses image derivatives for reducing ringing artifacts. Many techniques based on sparsity priors of image gradients have been proposed to deal with motion blur. Most previous methods assume that gradient magnitudes of natural images follow a heavy-tailed distribution. Fergus et al. \cite{FergusSingh} represent the heavy-tailed distribution over gradient magnitudes with a zero-mean mixture of Gaussian based on natural image statistics. Levin et al. \cite{LevinFergus} propose a hyper-Laplacian prior to fit the heavy-tailed distribution of natural image gradients. Shan et al. \cite{ShanJia} construct a natural gradient prior for the latent image by concatenating two piece-wise continuous convex functions.
However, sparse gradient priors always prefer the trivial solution, that is, the delta kernel and exactly the blurry image as the latent image estimate because the blur reduces the overall gradient magnitude. To tackle this problem, there are mainly two streams of research works for blind deconvolution. They use the maximum marginal probability estimation of $\boldsymbol{h}$ alone (marginalizing over $\boldsymbol{x}$) to recover the true kernel \cite{LevinWeiss2009,LevinWeiss2011,FergusSingh} or optimize directly the joint posterior probability of both $\boldsymbol{x}$ and $\boldsymbol{h}$ by performing some empirical strategies or heuristics to avoid the trivial solution during the minimization \cite{ShanJia,PerroneFavaro,PerroneFavaro2016}. Levin et al. \cite{LevinWeiss2009,LevinWeiss2011} suggest that a MAP (maximum a posterior) estimation of $\boldsymbol{h}$ alone is well conditioned and recovers an accurate kernel, while a simultaneous MAP estimation for solving blind deconvolution by jointly optimizing $\boldsymbol{x}$ and $\boldsymbol{h}$ would fail because it favors the trivial solution. 
Perrone and Favaro \cite{PerroneFavaro,PerroneFavaro2016} confirm the analysis of Levin et al. \cite{LevinWeiss2009,LevinWeiss2011} and conversely also declare that total
variation-based blind deconvolution methods can work well by performing specific implementation. In their work, the total variation regularization weight is initialized with a large value to help avoiding the trivial solution and iteratively reduced to allow for the recovery of more details. Blind deblurring is in general achieved through an alternating optimization scheme. In \cite{PerroneFavaro,PerroneFavaro2016}, the projected alternating minimization (PAM) algorithm of total variation blind deconvolution can successfully achieve the desired solution.

More present-day works often involve priors over larger neighborhoods or image patches, such as image super resolution \cite{PanYu}, image denoising \cite{WangYu}, no-blind image deblurring \cite{JiaEvans} and more. Gradient priors often consider two or three neighboring pixels, which are not sufficient for modeling larger image structures. Patch priors that consider larger neighborhoods (\emph{e.g.}, $5 \times 5$ or $7 \times 7$ image patches) model more complex structures and dependencies in larger neighborhoods. Image patches are usually overlapped with each other to suppress block effect. Sun et al. \cite{SunCho} use a patch prior learned from an external collection of sharp natural images to restore sharp edges. Michaeli and Irani \cite{MichaeliIrani} construct a cross-scale patch recurrence prior for the estimation of the blur kernel. Lai et al. \cite{LaiDing} obtain two color centers for every image patch and build a normalized color-line prior for blur kernel estimation. More recently, Pan et al. \cite{PanSun} introduce the dark channel prior based on statistics of image patches to kernel estimation, while Yan et al. \cite{YanRen} propose a patch-based bright channel prior for kernel estimation.

Recent work suggests that image patches can always be well represented sparsely with respect to an appropriate dictionary and the sparsity of image patches over the dictionary can be used as an image prior to regularize the ill-posed inverse problem. Zhang et al. \cite{ZhangYang} use sparse representation of image patches as a prior for blur kernel estimation and learn an over-complete dictionary from a collection of natural images or the observed blurry image itself using the K-SVD algorithm. Li et al. \cite{LiZhang} combine the dictionary pair and the sparse gradient prior with assumption that the blurry image and the sharp image have the same sparse coefficients under the blurry dictionary and the sharp dictionary respectively, to restore the sharp image via sparse reconstruction using the blurry image sparse coefficients on the sharp dictionary. The key issue of sparse representation is to identify a specific dictionary that best represents latent image patches in a sparse manner. Most methods use a database collecting enormous images as training samples to learn a universal dictionary. To make each patch of the latent image sparsely represented over such a universal dictionary, the database need involve massive training images, and thus this may lead to an inefficient learning and a potentially unstable dictionary. Meanwhile, the database needs to provide patches similar to the patches from the latent image, which cannot hold all the time. Alternatively, the dictionary is trained from the observed blurry image itself. However, the sparsity of the latent sharp image over the learned dictionary cannot be constantly guaranteed.

In this paper, we focus on an edge-based regularization approach for blind motion deblurring using patch priors. In our previous work, sparse representation and self-similarity are combined to work for image super resolution (SR) \cite{PanYu}. Super resolution approaches typically assume that the blur kernel is known (either the point spread function of the camera, or some default low-pass filter, \emph{e.g.} a Gaussian), while blind deblurring refers to the task of estimating the unknown blur kernel. Michaeli and Irani \cite{MichaeliIrani} have showed image super resolution approaches cannot be applied directly to blind deblurring. In \cite{YuChang}, we have proposed a blur kernel estimation method for blind motion deblurring using sparse representation and self-similarity of image patches as priors to guide the recovery of the latent image. In the previously proposed method, we construct a sparsity regularizer and a cross-scale non-local regularizer based on our priors. This method works quite well for a wide range of blurs but fails to deal with some extremely difficult cases. The edge-based method proposed in this paper is based on the observation that our regularizers often prefer sharp images to blurry ones only for image patches of salient edges. This fundamental observation enable us to build our regularizers on salient edge patches. Finally, we take an approximate iterative approach to solve the optimization problem by alternately updating the blur kernel and the latent image in a coarse-to-fine framework. 

The remainder of this paper is organized as follows. Section~\ref{sec:theory} describes the background on sparse representation and multi-scale self-similarity. Section~\ref{sec:method} makes detailed description on the proposed method, including our patch regularizers, our blind deconvolution model and the solution to our model. Section~\ref{sec:results} presents experimental results on both simulated and real blurry images. Section~\ref{sec:conclusion} draws the conclusion.
\section{SPARSE REPRESENTATION AND MULTI-SCALE SELF-SIMILARITY}
\label{sec:theory}
\subsection{Sparse Representation}
\label{ssec:sparserepresentation}
Image patches can always be represented well as a sparse linear combination of atoms (\emph{i.e.} columns) in an appropriate dictionary. Suppose that the image patch can be represented as $\mathbf{Q}_j\boldsymbol{X}$, here $\mathbf{Q}_j\in\mathbb{R}^{n\times N}$ is a matrix extracting the $ j $th patch from $\boldsymbol{X} \in \mathbb{R}^N$ ordered lexicographically by stacking either the rows or the columns of $\boldsymbol{x}$ into a vector, and the image patch $\mathbf{Q}_j\boldsymbol{X}\in\mathbb{R}^n$ can be represented sparsely over $\mathbf{D}\in\mathbb{R}^{n\times t}$, that is:
\begin{equation}
\mathbf{Q}_j\boldsymbol{X} = \mathbf{D}\boldsymbol{\alpha}_j,\Vert{\boldsymbol{\alpha}_j}\Vert_0 \ll n,
\label{eq:SR}
\end{equation}
where $\mathbf{D} =\left [\boldsymbol{d}_1,\cdots,\boldsymbol{d}_t\right] \in \mathbb{R}^{n\times t}$ refers to the dictionary, each column $\boldsymbol{d}_j \in \mathbb{R}^n $ for $j=1,\cdots,t $ represents the atom of the dictionary $\mathbf{D} $, ${\boldsymbol{\alpha}}_j =[\alpha_1,\cdots,\alpha_t]^{\rm T} \in \mathbb{R}^t $ is the sparse representation coefficient of $\mathbf{ Q}_j\boldsymbol{X}$ and $\Vert\boldsymbol{\alpha}_j\Vert_0$ counts the nonzero entries in $\boldsymbol{\alpha}_j$. 

Given a set of training samples $\boldsymbol{s}_i\in\mathbb{R}^{n},i=1,\cdots,m$, here $m$ is the number of training samples, dictionary learning attempts to find a dictionary $\mathbf{D} $ that forms sparse representations $\boldsymbol{\alpha}_i,i=1,\cdots,m$ for the training samples by jointly optimizing $\mathbf{D}$ and $\boldsymbol{\alpha}_i,i=1,\cdots,m$ as follows:
\begin{equation}
\min\limits_{\mathbf{D},\boldsymbol{\alpha}_1,\cdots,\boldsymbol{\alpha}_m}\sum\limits_{i=1}^{m}\Vert\boldsymbol{s}_i-\mathbf{D}\boldsymbol{\alpha}_i\Vert_2^2\quad {\rm{s.t.}}\ \forall i \ \Vert\boldsymbol{\alpha}_i\Vert_0 \leqslant T,
\label{eq:DictionaryLearning}
\end{equation}
where $T \ll n$ controls the sparsity of $\boldsymbol{\alpha}_i$ for $i=1,\cdots,m$. The K-SVD method \cite{AharonElad} is an effective dictionary learning method which solves Eq.(\ref{eq:DictionaryLearning}) by alternately optimizing $\mathbf{D}$ and $\boldsymbol{\alpha}_i,i=1,\cdots,m$. 

We firstly use the K-SVD method \cite{AharonElad} to obtain the dictionary $\mathbf{D}$. Then, we have to derive the sparse coefficient $\boldsymbol{\alpha}_j$ for the patch $\mathbf{ Q}_j\boldsymbol{X}$. Eq.(\ref{eq:SR}) can be formulated as the following $\ell_0$-norm minimization problem:
%\begin{equation}
%    \min\limits_{\boldsymbol{\alpha}_j} ||\boldsymbol{\alpha}_j||_0 \quad {\rm{s.t.}}\ %\Vert\mathbf{Q}_j\boldsymbol{X}-D\boldsymbol{\alpha}_j\Vert_2^2 \leq \epsilon
%\label{eq:SRCoeff}
%\end{equation}
\begin{equation}
    \min\limits_{\boldsymbol{\alpha}_j}\Vert\mathbf{ Q}_j{\boldsymbol{X}}-{\mathbf{D}}\boldsymbol{\alpha}_j\Vert_2^2\quad {\rm{s.t.}}\ \Vert{\boldsymbol{\alpha}}_j\Vert_0\leqslant T,
\label{eq:SRCoeffT}
\end{equation}
where $T$ is the sparsity constraint parameter. In our method, we obtain an approximation solution $\boldsymbol{\hat {\alpha}}_j$ for Eq.(\ref{eq:SRCoeffT}) by using the orthogonal matching pursuit (OMP) method \cite{TroppGilbert}. 

As a matter of fact, the precision of the K-SVD method can be controlled either by constraining the representation error or by constraining the number of nonzero entries in $\boldsymbol{\alpha}_i$. We use the latter formulated in Eq.(\ref{eq:DictionaryLearning}), because it is required in the OMP method \cite{TroppGilbert}. In other words, the objective could be met by constraining the number of nonzero entries in the sparse representation coefficients $\boldsymbol{\alpha}_i $. 
Once the sparse coefficient $\boldsymbol{\hat {\alpha}}_j$ is derived by solving Eq.(\ref{eq:SRCoeffT}), the reconstructed image patch $\mathbf {Q}_j\hat{\boldsymbol{X}}$ can be represented 
sparsely over $\mathbf{D}$ through $\mathbf{Q}_j \boldsymbol{\hat{X}} = \mathbf{D} {\boldsymbol {\hat\alpha}}_j$.

%
%where $\epsilon$ is a parameter that controls the representation error. According to (\ref{eq:SRCoeff}) the representation error is no more than $\epsilon$; at the same time, $\Vert \boldsymbol{\alpha}_i  \Vert _0 $ should be the minimum for every $j=1,\ldots,p $.
%
\subsection{Multi-Scale Self-Similarity and Non-local Regularization}
\label{ssec:multiscale}
Most natural images have properties of multi-scale self-similarity: structures from image fragments tend to repeat themselves at the same or different scales in natural images. In particular when small image patches are used, patch repetitions are found abundantly in multiple image scales of almost any natural image, even when we do not visually perceive any obvious repetitive structure. This is due to the fact that very small patches often contain only an edge, a corner, \emph{etc.} \cite{GlasnerBagon}. Glasner et al. \cite{GlasnerBagon} have showed that almost any image patch in a natural image has multiple similar patches in down-scaled versions of itself. 

\begin{figure}[htbp]
    \setlength{\fboxsep}{0cm}
    \setlength{\fboxrule}{0.6pt}    
    \centering
    \begin{subfigure}[Sharp image]
    {\begin{minipage}{0.4\textwidth}
        \centering
        \includegraphics[width=\textwidth]{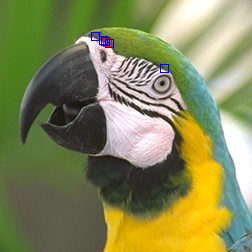}
        \\\vspace{1mm}
    \end{minipage}}
    \end{subfigure}
    \begin{subfigure}[Similar image patches within the same scale]
    {\begin{minipage}{0.4\textwidth}
        \centering
        \vspace{0.255\textwidth}
        \fcolorbox{red}{white}{\includegraphics[width=0.23\textwidth]{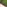}}
        \fcolorbox{blue}{white}{\includegraphics[width=0.23\textwidth]{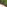}}
        \fcolorbox{blue}{white}{\includegraphics[width=0.23\textwidth]{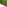}}
        \\\vspace{0.01\textwidth}
        \fcolorbox{blue}{white}{\includegraphics[width=0.23\textwidth]{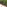}}
        \fcolorbox{blue}{white}{\includegraphics[width=0.23\textwidth]{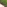}}
        \fcolorbox{blue}{white}{\includegraphics[width=0.23\textwidth]{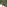}}
        \\
        \vspace{0.255\textwidth}
        \vspace{1mm}
    \end{minipage}}
    \end{subfigure}
    \\\vspace{0.03\textwidth}
    \begin{subfigure}[Down-sampled image]
    {\begin{minipage}{0.4\textwidth}
        \centering
        \includegraphics[width=0.5\textwidth]{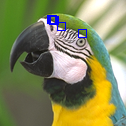}
        \\\vspace{1mm}
    \end{minipage}}
    \end{subfigure}
    \begin{subfigure}[Similar image patches across different scales]
    {\begin{minipage}{0.4\textwidth}
        \centering
        \vspace{0.01\textwidth}
        \fcolorbox{red}{white}{\includegraphics[width=0.23\textwidth]{1b1}}
        \fcolorbox{blue}{white}{\includegraphics[width=0.23\textwidth]{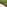}}
        \fcolorbox{blue}{white}{\includegraphics[width=0.23\textwidth]{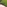}}
        \\\vspace{0.01\textwidth}
        \fcolorbox{blue}{white}{\includegraphics[width=0.23\textwidth]{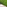}}
        \fcolorbox{blue}{white}{\includegraphics[width=0.23\textwidth]{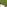}}
        \fcolorbox{blue}{white}{\includegraphics[width=0.23\textwidth]{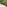}}
        \\\vspace{1mm}
    \end{minipage}}
    \end{subfigure}
    \caption{Patch repetitions occur abundantly both within the same scale and across different scales of a single image.}
    \label{fig:multi_scale_self-similarity}
\end{figure}

Fig.\ref{fig:multi_scale_self-similarity} schematically illustrates patch repetitions of self-similar structures both within the same scale and across different scales of a single image. For a patch of size $7 \times 7$ (marked with a red box) in Fig.\ref{fig:multi_scale_self-similarity}(a), we search for its $5$ similar patches (marked with blue boxes) in this image. Fig.\ref{fig:multi_scale_self-similarity}(b) shows close-ups of these similar patches within the same scale. In this example, the image is down-sampled by a factor of $a = 2$, as shown in Fig.\ref{fig:multi_scale_self-similarity}(c). For the patch marked with a red box in Fig.\ref{fig:multi_scale_self-similarity}(a) at the original scale, we also search for its $5$ similar patches of the same size in Fig.\ref{fig:multi_scale_self-similarity}(c), marked with blue boxes. Fig.\ref{fig:multi_scale_self-similarity}(d) shows close-ups of these similar patches searched from the down-sampled image, \emph{i.e.} cross-scale similar patches. The patches shown in Fig.\ref{fig:multi_scale_self-similarity} are displayed with clear repetitive structure in this image. 

%Peyre et. al propose a non-local regularization method for general inverse problems based on self-similarity properties of natural images \cite{PeyreBougleux}. 
The non-local means was firstly introduced for image denoising based on this self-similarity property of natural images in the seminal work of Buades \cite{BuadesColl}, and since then, the non-local means is extended succesfully to other inverse problems such as image super resolution and non-blind image deblurring \cite{ProtterElad,DongZhang}. 
%Recently, this self-similarity property is thoroughly explored by Glasner et. al in \cite{GlasnerBagon} for addressing super-resolution problems. 
The non-local means is based on the observation that similar image patches within the same scale are likely to be appeared in a single image, and these same-scale similar patches can provide additional information.
In our blind deconvolution model, we use similar image patches across different scales to construct a cross-scale non-local regularization prior by exploiting the correspondence between these cross-scale similar patches of the same image. Suppose that $\boldsymbol{X}\in\mathbb{R}^{N}$ and $\boldsymbol{X}^{a}\in\mathbb{R}^{N/a^2}$ represent the sharp image and its down-scaled version respectively, where $N$ is the size of the sharp image, and $a$ is the down-scaling factor. For each patch ${\mathbf Q}_j\boldsymbol{X}$ in the sharp image $\boldsymbol{X}$, we can search for its similar patches ${\mathbf R}_i\boldsymbol{X}^{a}$ in $\boldsymbol{X}^{a}$ that the similarity is measured by the distance between ${\mathbf Q}_j\boldsymbol{X}$ and ${\mathbf R}_i\boldsymbol{X}^{a}$, here ${\mathbf Q}_j\in\mathbb{R}^{n\times N}$ and ${\mathbf R}_i\in\mathbb{R}^{n\times N/a^2}$ are matrices extracting the $j$th and the $i$th patch from $\boldsymbol{X}$ and $\boldsymbol{X}^{a}$ respectively, and $n$ is the size of the image patch. The linear combination of the $L$ most similar patches of ${\mathbf Q}_j\boldsymbol{X}$ (put into the set $\mathcal{S}_j$) is used to predict ${\mathbf Q}_j\boldsymbol{X}$, that is, the prediction can be represented as the following weighted sum:
\begin{equation}
\mathbf{Q}_j\boldsymbol{X} \approx \sum_{i\in \mathcal{S}_j}w_i^j\mathbf{R}_i\boldsymbol{X}^{a},
\end{equation}
where
\begin{equation}
w_i^j=\frac{\exp(-\Vert\mathbf{Q}_j\boldsymbol{X}-\mathbf{R}_i\boldsymbol{X}^{a}\Vert_2^2/h)}{\sum_{l\in \mathcal{S}_j}\exp(-\Vert\mathbf{Q}_j\boldsymbol{X}-\mathbf{R}_l\boldsymbol{X}^{a}\Vert_2^2/h)}
\end{equation}
is the weight and $h$ is the control parameter of the weight. It is noted from self-similarity that any patch can, in some way, be approximated by other similar patches of the same image. Obviously the difference between $ \mathbf{Q}_j\boldsymbol{X} $ and its prediction should be small and the prediction error can be used as the regularization in our blind deconvolution model (\emph{i.e.} the cross-scale non-local regularizer).
\section{Blind Deconvolution}
\label{sec:method}
\subsection{Use of Cross-Scale Self-Similarity}
\label{ssec:cross-scale}
We incorporate both sparse representation and self-similarity of image patches as priors into our blind deconvolution model to regularize the recovery of the latent image with these priors as regularizers. Since patches repeat across scales in natural images, our patch-based regularizers can depend on abundant patch repetitions across different scales of the same image. Typically we partition the latent image into small overlapping patches. For every patch of the latent image, we search for similar patches of the same size in a down-scaled version of itself. We construct a sparsity regularizer by sparsely representing the latent sharp image over the dictionary that these cross-scale similar patches are used as training samples to learn, denoted by ${\rm Reg}_c(\boldsymbol{x})$:
\begin{equation}
     {\rm Reg}_c(\boldsymbol{x}) = \sum\limits_{j} \Vert {\mathbf Q}_j\boldsymbol{X}-{\mathbf D} \boldsymbol{\alpha}_j\Vert_2^2,
     \label{eq:sparse_reg} \\ 
\end{equation}  
and a cross-scale non-local regularizer according to the correspondence between the latent image patch and its similar patches searched from the down-scaled latent image to enforce the recovery of sharp edges, denoted by ${\rm Reg}_s(\boldsymbol{x})$:
\begin{equation}  
     {\rm Reg}_s(\boldsymbol{x}) = \sum\limits_j\Vert {\mathbf Q}_j\boldsymbol{X}-\sum\limits_{i\in\mathcal{S}_j}w_i^j {\mathbf R}_i\boldsymbol{X}^{a}\Vert_2^2,
     \label{eq:ssim_reg} 
\end{equation}
where $\mathbf{D}$ is the learned dictionary for sparse representation, $\boldsymbol{X}$ is the vector-form notion of $\boldsymbol{x}$, $\boldsymbol{X}^{a}$ is the down-scaled version of $\boldsymbol{X}$ by a factor $a$, ${\mathbf Q}_j\boldsymbol{X}$ and ${\mathbf R}_i\boldsymbol{X}^{a}$ represent the $j$th and the $i$th patch extracted from the latent image $\boldsymbol{X}$ and its down-scaled version $\boldsymbol{X}^{a}$ respectively, and $\mathcal{S}_j$ denotes the set of the $p$ most similar patches of ${\mathbf Q}_j\boldsymbol{X}$ searched from $\boldsymbol{X}^{a}$. We only use similar image patches at down-sampled scales of the latent image to construct the non-local regularizer, without involving those within the same scale into our non-local regularizer.

The choice of training samples is very important for dictionary learning problem. Ideally the dictionary $\mathbf{D}$ should be trained from the patches sampled from the unknown latent sharp image. In our previous single-image super-resolution work \cite{PanYu}, the dictionary is trained from the low-resolution image itself. Unforturnately, it is not a good choice for blind deblurring to learn a dictionary using the observed blurry image itself as training samples. This is because the dictionary trained from the blurry image cannot guarantee the sparsity of sharp image patches. In the previously proposed method \cite{YuChang}, we used an adaptive over-complete dictionary trained from the down-scaled blurry image, more similar to the latent sharp image than the blurry image itself. In this paper, we present an improvement to collect training samples from the down-scaled latent image estimate, as will be detailed later.

\begin{figure}[htbp]
    \setlength{\fboxsep}{0cm}
    \setlength{\fboxrule}{0.6pt}    
    \centering
    \begin{subfigure}[Blurry image]
    {\begin{minipage}{0.4\textwidth}
        \centering
        \includegraphics[width=\textwidth]{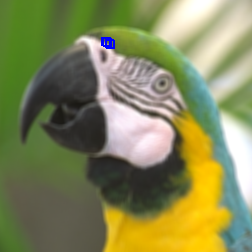}
        \\\vspace{1mm}
    \end{minipage}}
    \end{subfigure}
    \begin{subfigure}[Similar patches in blurry image]
    {\begin{minipage}{0.4\textwidth}
        \centering
        \vspace{0.255\textwidth}
        \fcolorbox{red}{white}{\includegraphics[width=0.23\textwidth]{1b1}}
        \fcolorbox{blue}{white}{\includegraphics[width=0.23\textwidth]{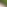}}
        \fcolorbox{blue}{white}{\includegraphics[width=0.23\textwidth]{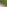}}
        \\\vspace{0.01\textwidth}
        \fcolorbox{blue}{white}{\includegraphics[width=0.23\textwidth]{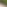}}
        \fcolorbox{blue}{white}{\includegraphics[width=0.23\textwidth]{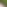}}
        \fcolorbox{blue}{white}{\includegraphics[width=0.23\textwidth]{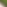}}
        \\
        \vspace{0.255\textwidth}
        \vspace{1mm}
    \end{minipage}}
    \end{subfigure}
    \\\vspace{0.03\textwidth}
    \begin{subfigure}[Down-sampled blurry image]
    {\begin{minipage}{0.4\textwidth}
        \centering
        \includegraphics[width=0.5\textwidth]{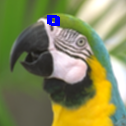}
        \\\vspace{1mm}
    \end{minipage}}
    \end{subfigure}
    \begin{subfigure}[Similar patches in down-sampled blurry image]
    {\begin{minipage}{0.4\textwidth}
        \centering
        \vspace{0.01\textwidth}
        \fcolorbox{red}{white}{\includegraphics[width=0.23\textwidth]{1b1}}
        \fcolorbox{blue}{white}{\includegraphics[width=0.23\textwidth]{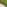}}
        \fcolorbox{blue}{white}{\includegraphics[width=0.23\textwidth]{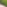}}
        \\\vspace{0.01\textwidth}
        \fcolorbox{blue}{white}{\includegraphics[width=0.23\textwidth]{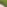}}
        \fcolorbox{blue}{white}{\includegraphics[width=0.23\textwidth]{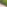}}
        \fcolorbox{blue}{white}{\includegraphics[width=0.23\textwidth]{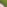}}
        \\\vspace{1mm}
    \end{minipage}}
    \end{subfigure}
    \caption{Down-scaled blurry patches are more similar to the sharp patch than blurry patches at the original scale.}
    \label{fig:self_similarity_sharp_blur}
\end{figure}

We now provide illustration to account for the use of cross-scale self-similarity. Although patches repeat within and across scales of the sharp image, as illustrated in Fig.\ref{fig:multi_scale_self-similarity}, the similarity diminishes significantly between the sharp image and its blurred counterpart. For the patch marked with a red box from the sharp image shown in Fig.\ref{fig:multi_scale_self-similarity}(a), we still search for its 5 most similar patches from the blurry image (Fig.\ref{fig:self_similarity_sharp_blur}(a)) and its down-scaled version (Fig.\ref{fig:self_similarity_sharp_blur}(c)) by using block matching, respectively. Fig.\ref{fig:self_similarity_sharp_blur} shows that the patches from the down-scaled blurry image (Fig.\ref{fig:self_similarity_sharp_blur}(d)) that are more similar to the patch from the sharp image than the patches from the blurry image itself (Fig.\ref{fig:self_similarity_sharp_blur}(b)). This is because the blur effect tends to weaken at coarser scales of the image despite the strong blur at the original scale. It is easy to verify that down-scaling an image by a factor of $a$ produces $a$-times sharper patches of the same size that are more similar to patches from the latent sharp image. Please refer to \cite{MichaeliIrani} for the proof.

\begin{figure}[htbp]
    \centering
    \includegraphics[width=0.99\columnwidth,clip=true]{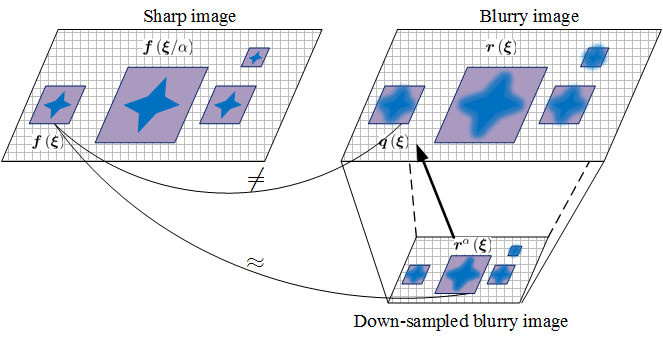}
%    \vskip1mm
    \caption{Similar patches across different scales are available for providing a prior for restoration.}
    \label{fig:self-similarity}
\end{figure}

Fig.\ref{fig:self-similarity} illustrates the reason why similar patches across different scales are available for providing a prior for restoration. Suppose that ${f}\left(\boldsymbol{\xi}\right)$ and ${f}\left (\boldsymbol{\xi}/a\right )$ are cross-scale similar patches and ${f}\left (\boldsymbol{\xi}/a\right )$ is an ${a}$-times larger patch in the sharp image, here $\boldsymbol{\xi}$ denotes the spatial coordinate. Accordingly, their blurry counterparts ${q}\left (\boldsymbol{\xi}\right )$ and ${r}\left (\boldsymbol{\xi}\right )$ are similar across image scales, and the size of ${r}\left (\boldsymbol{\xi}\right )$ is $a$ times as large as that of ${q}\left (\boldsymbol{\xi}\right )$ in the blurry image. In Fig.\ref{fig:self-similarity}, the blurry image is $a$ times the size of its down-sampled version. Down-scaling the blurry patch ${r}\left (\boldsymbol{\xi}\right )$ by a factor of ${a}$ generates an ${a}$-times smaller patch ${{r}}^{a}\left (\boldsymbol{\xi}\right )$. Obviously, ${q}\left (\boldsymbol{\xi}\right )$ and ${{r}}^{a}\left (\boldsymbol{\xi}\right )$ are of the same size and the patch ${{r}}^{a}\left (\boldsymbol{\xi}\right )$ from the down-sampled image is exactly an $a$-times sharper version of the patch ${{q}}\left (\boldsymbol{\xi}\right )$ in the blurry image. In such a case, ${{r}}^{a}\left (\boldsymbol{\xi}\right )$ can offer much exact prior information for the recovery of ${q}\left (\boldsymbol{\xi}\right )$. Fig.\ref{fig:self-similarity} schematically demonstrates that the patches at coarser image scales can serve as a good prior, although it is an ideal case.

In summary, we incorporate effectively prior knowledge provided by cross-scale similar patches into our regularizers. As stated above, the down-scaled latent image estimate can provide sharper patches of the same size that are more similar to patches from the latent sharp image. In the sparsity regularizer, the dictionary is trained from sharper patches sampled from the down-scaled latent image estimate to make latent image patches well represented sparsely. In the cross-scale non-local regularizer, meanwhile, all latent image patches are optimized to be as close to their sharper similar patches searched from the down-scaled latent image estimate to enforce the sharp recovery of the latent image as possible.
\subsection{Analysis on Regularizers}
\label{ssec:priors}
In regularization approaches, blind deconvolution is generally formulated as an energy minimization problem with appropriate regularizers, which tends to be minimal at the desired latent image. The regularizers are used to impose additional constraints on the optimization problem. They significantly benefit the solution of the blind deconvolution problem based on the condition that the regularization functions with respect to the sharp image $\boldsymbol{x}$ should be significantly smaller than those with respect to its blurry counterpart $\boldsymbol{y}$. We will make the sparsity and the self-similarity comparison between the sharp image and the blurry image based on our patch regularizers respectively, and discuss whether the condition is satisfied or which patches satisfy this condition.

\subsubsection{Sparsity Regularizer}
\label{ssec:sparisty}
First of all, we compare the sparsity regularization functions ${{\rm Reg}_c}(\boldsymbol{x})$ and ${{\rm Reg}_c}(\boldsymbol{y})$ with respect to the sharp image $\boldsymbol{x}$ and the blurry one $\boldsymbol{y}$, respectively. For comparison, we generate the blurred image by the convolution of the sharp image shown in Fig.\ref{fig:multi_scale_self-similarity}(a) with the averaging blur kernel. The dictionary is trained from patches sampled from the down-sampled blurry image. We calculate the values of the sparsity regularization functions with respect to the sharp image and several blurred images with blur kernels of varying sizes of $ 2\times 2 $, $ 3\times 3 $ and $ 5\times 5 $, respectively, which are averaged over all pixels, as shown in Table \ref{tab:sparsity_x_y}, where $ N $ is the size of the image, $ n $ is the size of image patch. The smaller the value, the smaller the sparse representation error. This means that the image is better represented over the learned dictionary. From Table \ref{tab:sparsity_x_y}, we can see that the sharp image has larger sparse representation error than any blurred image over the learned dictionary, and the larger blur corresponds to the sparser representation of the blurred image in terms of the entire image.

\begin{table}[htbp]
	\centering
	\caption{Comparison of sparsity regularizer between sharp image and blurry images with blur kernels of different sizes}
    \label{tab:sparsity_x_y}
    \vskip1mm
    \begin{minipage}[t]{0.9\textwidth}
    \renewcommand{\baselinestretch}{1.35}
    \begin{tabular}{ccccc}
    	\toprule[1pt]
    		     &  Sharp  & $2\times 2$ blur     &  $3\times 3$ blur    & $5\times 5$ blur\\ 
    	\midrule[0.5pt]
    	$\sqrt{{\rm Reg}_c(\cdot)/(N\cdot n)}$				 &	5.40	& 				3.70				      &     2.70	                           &     1.72           \\
    	\bottomrule[1pt]
    \end{tabular}
    \vspace{1mm}
    \\\footnotesize Note: the intensity range is $[0,1]$.
    \end{minipage}
\end{table}

Then we compare the sparsity regularization functions with respect to the sharp image and the blurred counterpart
on a patch-by-patch basis. Let $\mathcal{R}_{c}$ represent the set of pixels at which the sharp patch has smaller sparse representation error than the blurred one over the learned dictionary. That is,
\begin{equation}
\mathcal{R}_c=\{j|\ \Vert {\mathbf Q}_j\boldsymbol{X}-{\mathbf D}\boldsymbol{\alpha}_j\Vert_2^2\leqslant\Vert {\mathbf Q}_j\boldsymbol{Y}-{\mathbf D}\boldsymbol{\alpha}_j\Vert_2^2\},
\end{equation}
where $\boldsymbol{X}$ and $\boldsymbol{Y}$ denote the vector notations of the sharp image $\boldsymbol{x}$ and the blurred image $\boldsymbol{y}$ respectively. Fig.\ref{fig:our_priors_on_edge}(a) shows the blurred image with the averaging blur kernel of size $2\times 2$. In Fig.\ref{fig:our_priors_on_edge}(b), the set $ \mathcal{R}_c $ are indicated with white pixels where the sharp patch achieves smaller sparse representation error than the blurred patch over the learned dictionary. From Fig.\ref{fig:our_priors_on_edge}(b), we can see that the sparsity regularizer of the sharp image is smaller than that of the blurred image only for some certain patches. 
%and the larger the blur, the larger the region. 
Intuitively, these regions comprised of white pixels coincide with edges and sharp changes in this image. It is believed that most image structures are often reflected around edges and areas of high variation. The optimal dictionary should produce sparsest representation of edge patches in the latent sharp image.

\subsubsection{Non-local Regularizer}
\label{ssec:selfsimilarity}

For the same reason, we compare the non-local regularization functions ${{\rm Reg}_s}(\boldsymbol{x})$ and ${{\rm Reg}_s}(\boldsymbol{y})$ with respect to the sharp image $\boldsymbol{x}$ and the blurry one $\boldsymbol{y}$, respectively. Similarly, we calculate the values of the non-local regularization functions with respect to the sharp image and the blurred images with blur kernels of varying sizes of $ 2\times 2 $, $ 3\times 3 $ and $ 5\times 5 $ respectively, averaged over all pixels, as shown in Table \ref{tab:ssim_x_y}. The smaller the value, the smaller the prediction error. It means that there is stronger cross-scale self-similarity throughout the image. From Table \ref{tab:ssim_x_y}, we can see that the sharp image reveals the weakest cross-scale self-similarity, and the blurred image with larger blur kernel displays stronger cross-scale self-similarity in terms of the entire image.

\begin{table}[htbp]
	\centering
	\caption{Comparison of cross-scale non-local regularizer between sharp image and blurry images with blur kernels of different sizes}
    \label{tab:ssim_x_y}
    \vskip1mm
    \begin{minipage}[t]{0.9\textwidth}
        \renewcommand{\baselinestretch}{1.35}
    	\begin{tabular}{ccccc}
		    \toprule[1pt]
                                        & Sharp  & $2\times 2$ blur  & $3\times 3$ blur  & $5\times 5$ blur\\
		    \midrule[0.5pt]
            $\sqrt{{\rm Reg}_s(\cdot)/(N\cdot n)}$    &   0.0448     &  0.0385                & 0.0339                  & 0.0271  \\
		    \bottomrule[1pt]
    	\end{tabular}
    \vspace{1mm}
    \\\footnotesize Note: the intensity range is $[0,1]$.
    \end{minipage}
\end{table}

\begin{figure}[htbp]
    \centering
        \subfigure[{Blurred image}]
    {\begin{minipage}{0.4\textwidth}
        \centering
        \includegraphics[width=\textwidth]{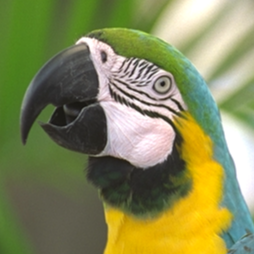}
        \vspace{0.5mm}
    \end{minipage}}
    \subfigure[{$ \mathcal{R}_c $}]
    {\begin{minipage}{0.4\textwidth}
        \centering
        \includegraphics[width=\textwidth]{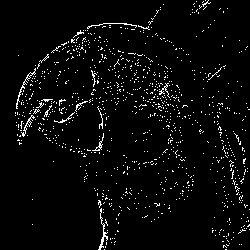}
        \vspace{0.5mm}
    \end{minipage}}
        \subfigure[{$ \mathcal{R}_s $}]
    {\begin{minipage}{0.4\textwidth}
        \centering
        \includegraphics[width=\textwidth]{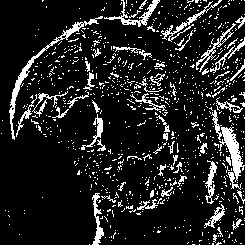}
        \vspace{0.5mm}
    \end{minipage}}
    \subfigure[{Salient edges of (a)}]
    {\begin{minipage}{0.4\textwidth}
        \centering
        \includegraphics[width=\textwidth]{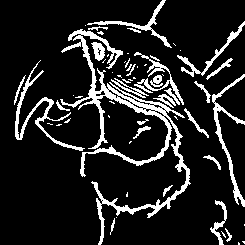}
        \vspace{0.5mm}
    \end{minipage}}
    \caption{Sharp image has stronger sparsity and cross-scale self-similarity than blurred image roughly around salient edges.}
    \label{fig:our_priors_on_edge}
\end{figure}

We still compare the non-local regularization functions with respect to the sharp image and the blurred counterpart on a patch-by-patch basis. Let $\mathcal{R}_{s}$ represent the set of pixels at which the sharp patch has smaller prediction error than the blurred one. That is,
\begin{equation}
\mathcal{R}_s=\{j|\ \Vert {\mathbf Q}_j\boldsymbol{X}-\sum_{i\in\mathcal{S}_j}w_i^j{\mathbf R}_i\boldsymbol{X}^{a}\Vert_2^2\leqslant\Vert {\mathbf Q}_j\boldsymbol{Y}-\sum_{i\in\mathcal{S}_j}w_i^j {\mathbf R}_i\boldsymbol{Y}^{a}\Vert_2^2\},
\end{equation}
where $\boldsymbol{Y}$ and $\boldsymbol{Y}^{a}$ denote the vector notation of the blurred image $\boldsymbol{y}$ and its down-sampled version by a factor of $a$. From Fig.\ref{fig:our_priors_on_edge}(c), the set $ \mathcal{R}_s $ indicated with white pixels is also roughly consistent with edges of the image.
Our further observation shows that image edges do not always help kernel estimation when the scale of the edge is smaller than that of the blur kernel, while salient edges can effectively avoid the trivial solution and get an accurate blur kernel. 
%To verify this observation, 
We use Sun et al's strategy \cite{SunCho} (see the following edge mask ${\boldsymbol {M}}$ for more details) to detect and select salient edges of the blurred image, as is shown in Fig.\ref{fig:our_priors_on_edge}(d). 
%It can been observed that the sparsity of the sharp image is stronger than that of the blurred image around edges. 

It can be observed from the comparison of Figs. \ref{fig:our_priors_on_edge}(c) and (d) that the cross-scale non-local regularizer of the sharp image is smaller than that of the blurred image roughly around salient edges. The blur alters to different extent edges of repetitive structures across different scales and thus deteriorates cross-scale self-similarity properties of edge structures in the blurry image. 

%\begin{figure}[htbp]
%    \centering
%%    \subfigure[{Blurred image with averaging blur kernel of size $2\times 2$}]
%%    {\begin{minipage}{0.24\textwidth}
%%        \centering
%%        \includegraphics[width=\textwidth]{y_2x2}
%%        \vspace{1mm}
%%    \end{minipage}}
%    \subfigure[{Stronger self-similarity regions in sharp image than in the $2\times 2$ averaging blurred image}]
%    {\begin{minipage}{0.24\textwidth}
%        \centering
%        \includegraphics[width=\textwidth]{true_mask_sim_2x2}
%        \vspace{1mm}
%    \end{minipage}}
%    %\\\vspace{0.03\textwidth}
%%    \subfigure[{Blurred image with averaging blur kernel of size $3\times 3$}]
%%    {\begin{minipage}{0.24\textwidth}
%%        \centering
%%        \includegraphics[width=\textwidth]{y_3x3}
%%        \vspace{1mm}
%%    \end{minipage}}
%    \subfigure[{Stronger self-similarity regions in sharp image than in the $3\times 3$ averaging blurred image}]
%    {\begin{minipage}{0.24\textwidth}
%        \centering
%        \includegraphics[width=\textwidth]{true_mask_sim_3x3}
%        \vspace{1mm}
%    \end{minipage}}
%    \caption{Cross-scale self-similarity of the sharp image is stronger than that of the blurred image only around edges and sharp changes.}
%    \label{fig:ssim_prior_on_edge}
%\end{figure}

\subsection{Modeling and Optimization}
\label{ssec:optimization}
Although natural images generally have properties of sparsity and self-similarity, in the previous part, we have made detailed discussions on our two regularizers ${\rm Reg}_c(\boldsymbol{x})$ and ${\rm Reg}_s(\boldsymbol{x})$, and come to the conclusion that ${\rm Reg_c}(\boldsymbol{x})<{\rm Reg_c}(\boldsymbol{y})$ and ${\rm Reg_s}(\boldsymbol{x})<{\rm Reg_s}(\boldsymbol{y})$ are often satisfied only for image patches of salient edges. In other words, they only favor the sharp solution over the blurred one around salient image edges. In order to generate more exact solutions, our regularization constraints are only imposed on image patches of salient edges. 

In this paper, we define the edge mask ${\boldsymbol {M}}$ according to the corresponding salient edge pixels, which is a binary mask indicating pixel locations that we want to apply our priors. We employ a heuristic process to detect and select salient edges of the latent image estimate during the optimization in a coarse-to-fine framework for kernel estimation and thus we do not present a joint energy minimization formulation of both the latent image $\boldsymbol{x}$ and the blur kernel $\boldsymbol{h}$. In each level of the image pyramid, we take an approximate approach to solve the optimization problem by directly alternating between optimizing the kernel $\boldsymbol{h}$ and the latent image $\boldsymbol{x}$.

{\textbf {1. Updating ${\boldsymbol {M}}$}}

This step chooses pixel locations to apply our patch priors. Since our regularizers prefer the sharp image to the blurry one only around salient edges, in order to benefit the blur kernel estimation, we first detect and select useful salient edges. We adopt Sun et al.'s strategy \cite{SunCho} to filter the latent image estimate ${\boldsymbol{\hat x}}_{k}$ with a filter bank consisting of derivatives of Gaussians in eight directions and obtain the edge mask ${\boldsymbol {M}}$ by keeping the top $ 2\% $ of pixel locations from the largest filter responses of the filter bank. In our model, regions outside the mask are weakly regularized by our patch priors, resulting in noise amplification in flat or smooth regions, and therefore the Gaussian low-pass filter are utilized before salient edge selection.

{\textbf {2. Updating $\boldsymbol{h}$}}

In this step, we fix ${\boldsymbol{\hat x}}_k$ and update ${\boldsymbol{\hat h}}_{k+1}$. The minimization problem is defined with a Gaussian regularizer as:
\begin{equation}
     \hat{\boldsymbol{h}}_{k+1}=\arg\min\limits_{\boldsymbol{h}}  \Big\{\Vert\nabla\boldsymbol{y} - \boldsymbol{h} *  (\nabla\hat{\boldsymbol{x}}_k \odot {\boldsymbol {M}}) \Vert_2^2 + \lambda_h\Vert\boldsymbol{h}\Vert_2^2 \Big\},
     \label{eq:ObjFuncKEdge}
\end{equation}
where $\nabla=\{\partial_{x},\partial_{y}\}$ denotes the spatial derivative operator in two directions, $\odot$ stands for the pixel-wise multiplication, and $\lambda_h$ is the regularization weight to control the tradeoff between the fidelity to the observation model (as accounted for by the former term)
and the smoothness of the estimated blur kernel (as reflected by the latter term). We multiply $\nabla\hat{\boldsymbol{x}}_k$ by the mask ${\boldsymbol {M}}$ (\emph{i.e.} $\nabla\hat{\boldsymbol{x}}_k\odot {\mathbf {M}}$) to enforce that regions outside the mask do not participate in estimating $\boldsymbol{h}$. We only allow salient edges in the mask ${\boldsymbol {M}}$ to participate in the constraint of the observation model by setting the gradient $\nabla\hat{\boldsymbol{x}}_k$ outside ${\boldsymbol {M}}$ to zero. 

On the other hand, we take a common way to eliminate the influence of smooth or flat regions of the image on kernel estimation \cite{ChoLee,XuJia,SunCho,LaiDing}. The pixels whose gradient magnitudes are less than a certain threshold in the intermediate latent image estimate are set to zero. Let $\tau $ denote a threshold of the gradient magnitude and $N_h$ denote the size of the blur kernel. The threshold for truncating gradients is determined as follows. We construct the histograms of gradient magnitudes and directions for each $ \partial_*{\boldsymbol{ \hat x}}_k $.  Angles are quantized by $ 45^{\circ} $, and gradients of opposite directions are counted together. Then, we find a threshold that keeps at least $r\sqrt{N_h}$ pixels from the largest magnitude for each quantized angle. We use 2 for $ r $ by default. To allow for inferring subtle structures during kernel refinement, we gradually decrease the value of the threshold $\tau$ in iterations by dividing by $ 1.1 $ at each iteration, to include more and more edges.
Eq.(\ref{eq:ObjFuncKEdge}) excludes part of the gradients, depending jointly on the magnitude and the edge mask ${\boldsymbol {M}}$. In order to suppress the noise in flat or smooth regions, however, we do nothing on $ \nabla\boldsymbol{y} $. This selection process reduces ambiguity in the following kernel estimation.

Eq.(\ref{eq:ObjFuncKEdge}) is a quadratic funciton of unknown $\boldsymbol{h}$, which has a closed-form solution for $\hat{\boldsymbol{h}}_{k+1}$. We solve Eq.(\ref{eq:ObjFuncKEdge}) in the Fourier domain by performing FFTs on all variables and setting the derivative with respect to $\boldsymbol{h}$ to zero:
\begin{equation}
    \hat{\boldsymbol{h}}_{k+1} = \mathcal{F}^{-1}\left(\frac{\overline{\mathcal{F}(\partial_x{\boldsymbol{ \hat x}}_k\odot {\boldsymbol {M}})}\mathcal{F}(\partial_x\boldsymbol{y}) + \overline{\mathcal{F}(\partial_y{\boldsymbol{\hat x}}_k\odot {\boldsymbol {M}})}\mathcal{F}(\partial_y\boldsymbol{y})}{\mathcal{F}(\partial_x{\boldsymbol{\hat x}}_k\odot {\boldsymbol {M}})^2 + \mathcal{F}(\partial_y{\boldsymbol{\hat x}}_k\odot {\boldsymbol {M}})^2 + \lambda_h}\right),
    \label{eq:ObjFuncSolveKEdge}
\end{equation}
where $\mathcal{F}(\cdot)$ and $\mathcal{F}^{-1}(\cdot)$ denote the fast Fourier transform and inverse Fourier transform respectively, and $\overline{\mathcal{F}(\cdot)}$ is the complex conjugate operator.

{\textbf {3. Updating $\boldsymbol{x}$}}

In this step, we fix $\hat{\boldsymbol{h}}_{k+1}$, and given $\hat{\boldsymbol{x}}_k$ update $\hat{\boldsymbol{x}}_{k+1}$. With our patch priors as regularizers, we establish our regularizers on salient edge patches of the image, and  get the following regularized minimization:
\begin{equation}
    \begin{aligned}
        \hat{\boldsymbol{x}}_{k+1} = &\arg\min\limits_{\boldsymbol{x}}   \Big\{\Vert\nabla\boldsymbol{y}- \hat{\boldsymbol{h}}_{k+1} *  \nabla\boldsymbol{x} \Vert_2^2
         + \lambda_c\frac{N}{|{{\boldsymbol {M}}}|}\sum\limits_{j\in { {\boldsymbol {M}}}} \Vert{\mathbf Q}_j\boldsymbol{X}- {\mathbf D}\boldsymbol{\alpha}_j\Vert_2^2 \\
        & + \lambda_s\frac{N}{|{{\boldsymbol {M}}}|}\sum\limits_{j\in {{\boldsymbol {M}}}}\Vert{\mathbf Q}_j\boldsymbol{X}-\sum\limits_{i\in\mathcal{S}_j}w_i^j {\mathbf R}_i\boldsymbol{X}^{a}\Vert_2^2 
        + \lambda_g\Vert\nabla\boldsymbol{x}\Vert_2^2 \Big\}\\
        & {\rm{s.t.}}\ \forall j\ \Vert\boldsymbol{\alpha}_j\Vert_0\leqslant T
    \end{aligned},
    \label{eq:ObjFuncXEdge}
\end{equation}
where $\left|{\boldsymbol {M}}\right|$ is the number of non-zero elements in the mask ${\boldsymbol {M}}$, and $N$ is the size of the latent image, $\mathbf{D}$ is the dictionary trained from the down-scaled latent image estimate, $\boldsymbol{X}$ is the vector notation of the latent image $\boldsymbol{x}$, $\boldsymbol{X}^{a}$ is the down-sampled version of $\boldsymbol{X}$ by a factor of $a$, and $\lambda_c$, $\lambda_s$, and $\lambda_g$  are regularization weights controlling the effect of the regularizers. In Eq.(\ref{eq:ObjFuncXEdge}), the first term is the fidelity to the observation model, the second term is the sparsity regularizer, the third term is the cross-scale non-local regularizer, and the last term is the smoothness constraint of the estimated latent image.

Rearranging $\boldsymbol{y}$ in vector form, denoted by $\boldsymbol{Y}\in\mathbb{R}^{N}$, and rewriting the convolution of the blur kernel and the latent image in matrix-vector form, Eq.(\ref{eq:ObjFuncXEdge}) can be rewritten as
\begin{equation}
    \begin{aligned}
        \hat{\boldsymbol{X}}_{k+1} = & \arg\min\limits_{\boldsymbol{X}} \Big\{\Vert {\mathbf G}_x\boldsymbol{Y}-{\mathbf H}_{k+1} {\mathbf G}_x\boldsymbol{X}\Vert_2^2
        + \Vert {\mathbf G}_y\boldsymbol{Y}-{\mathbf H}_{k+1} {\mathbf G}_y\boldsymbol{X}\Vert_2^2\\
        & + \lambda_c\frac{N}{|{\boldsymbol M}|}\sum\limits_{j\in {\boldsymbol M}}\Vert {\mathbf Q}_j\boldsymbol{X}-{\mathbf D}\boldsymbol{\alpha}_j\Vert_2^2
         + \lambda_s\frac{N}{|{\boldsymbol M}|}\sum\limits_{j\in {\boldsymbol M}}\Vert {\mathbf Q}_j\boldsymbol{X}-\sum\limits_{i\in\mathcal{S}_j}w_i^j {\mathbf R}_i\boldsymbol{X}^{a}\Vert_2^2\\
          & + \lambda_g(\Vert {\mathbf G}_x\boldsymbol{X}\Vert_2^2+\Vert {\mathbf G}_y\boldsymbol{X}\Vert_2^2)\Big\}\\
         & {\rm{s.t.}}\ \forall j\ \Vert\boldsymbol{\alpha}_j\Vert_0\leqslant T
    \end{aligned},
    \label{eq:ObjFuncXVectorEdge}
\end{equation}
where $\mathbf{G}_x$ and $\mathbf{G}_y\in\mathbb{R}^{N\times N}$ are the matrix forms of the partial derivative operators $\partial_{x}$ and $\partial_{y}$ in two directions respectively, and $\mathbf{H}_{k+1}\in\mathbb{R}^{N\times N}$ is the blur matrix. Setting the derivative of Eq.(\ref{eq:ObjFuncXVectorEdge}) with respect to $\boldsymbol{X}$ to zero and letting $\mathbf{G}=\mathbf{G}_x^{\rm T}\mathbf{G}_x + \mathbf{G}_y^{\rm T}\mathbf{G}_y$, we derive
\begin{equation}
    \begin{array}{l}
        \big[({\mathbf H}_{k+1}^{\rm T}{\mathbf H}_{k+1}+\lambda_g){\mathbf G} + (\lambda_c+\lambda_s)\frac{N}{|{\boldsymbol {M}}|}\sum\limits_{j\in {\boldsymbol {M}}}{\mathbf Q}_j^{\rm T}{\mathbf Q}_j\big]\hat{\boldsymbol{X}}_{k+1}=\\
         {\mathbf H}_{k+1}^{\rm T} {\mathbf G}\boldsymbol{Y} + \lambda_c\frac{N}{|{\boldsymbol {M}}|}\sum\limits_{j\in {\boldsymbol {M}}} {\mathbf Q}_j^{\rm T} {\mathbf D}\boldsymbol{\alpha}_j 
        + \lambda_s\frac{N}{|{\boldsymbol {M}}|}\sum\limits_{j\in {\boldsymbol {M}}}{\mathbf Q}_j^{\rm T}\sum\limits_{i\in\mathcal{S}_j}w_i^j {\mathbf R}_i\hat{\boldsymbol{X}}^{a}_{k+1}
    \end{array},
\label{eq:ObjFuncXDerivativeEdge}
\end{equation}
Since both sparse representation coefficients $\boldsymbol{\alpha}_j$ and the down-sampled image $\boldsymbol{\hat X}^{a}_{k+1}$ on the right-hand side of Eq.(\ref{eq:ObjFuncXDerivativeEdge}) depend on unknown $\boldsymbol{\hat X}_{k+1}$, Eq.(\ref{eq:ObjFuncXDerivativeEdge}) cannot be solved in closed form. Instead we approximately solve Eq.(\ref{eq:ObjFuncXDerivativeEdge}) with the following procedure:

(1) The K-SVD method \cite{AharonElad} is used to attain the dictionary $ {\mathbf D} $ by approximately solving Eq.(\ref{eq:DictionaryLearning}). For each patch $\mathbf{Q}_j{\boldsymbol{\hat X}}_k$ in ${\boldsymbol{\hat X}}_k$ that the mask ${\boldsymbol {M}}$ selects, the OMP method \cite{TroppGilbert} is used here to derive the sparse representation coefficient $\boldsymbol{\alpha}_j$ over the dictionary $\mathbf{D}$ by approximately solving the following constrained minimization problem:
\begin{equation}
    {\boldsymbol{\hat \alpha}}_j=\arg\min\limits_{\boldsymbol{\alpha}_j}\Vert {\mathbf Q}_j {\boldsymbol{\hat X}}_{k}-{\mathbf D}\boldsymbol{\alpha}_j\Vert_2^2\quad {\rm {s.t.}}\ \Vert\boldsymbol{\alpha}_j\Vert_0\leqslant T.
\end{equation}
Since the sparse coefficient $\boldsymbol{\alpha}_j$ on the right-hand side of Eq.(\ref{eq:ObjFuncXDerivativeEdge}) depends on unknown $\boldsymbol{\hat X}_{k+1}$, we approximate ${\boldsymbol{\hat X}}_{k+1}$ using ${\boldsymbol{\hat X}}_k$ to solve the sparse coefficient $ {\boldsymbol{\hat \alpha}}_j $ over the dictionary $\mathbf{D}$.

(2) For the same reason, since ${\boldsymbol{\hat X}}_{k+1}$ and its down-scaled ${\boldsymbol{\hat X}}^{a}_{k+1}$ are both unknown, we approximate ${\boldsymbol{\hat X}}_{k+1}$ and ${\boldsymbol{\hat X}}^{a}_{k+1}$ using ${\boldsymbol{\hat X}}_{k}$ and $ {\boldsymbol{\hat X}}^{a}_{k} $ respectively. For each patch $\mathbf{Q}_j{\boldsymbol{\hat X}}_k$ in $\hat{\boldsymbol{X}}_k$ that the mask ${\boldsymbol {M}}$ selects, we search for its similar patches $\mathbf{R}_i {\boldsymbol{\hat X}}_k^{a}$, $ i\in {\hat {\mathcal{S}}}_j $ in the down-scaled image $\hat{\boldsymbol{X}}_k^{a}$ of $\hat{\boldsymbol{X}}_k$, and use the linear combination of these similar patches $\sum_{i\in {\hat {\mathcal{S}}}_j}{\hat w_i}^j\mathbf{R}_i {\boldsymbol{\hat X}}_k^{a}$ to predict it. Here $ {\hat {\mathcal{S}}}_j $ and $ {\hat w_i}^j $ are updated according to ${\boldsymbol{\hat X}}_{k}$ and $ {\boldsymbol{\hat X}}^{a}_{k} $.

(3) Eq.(\ref{eq:ObjFuncXDerivativeEdge}) can be reformulated by substituting the sparse coefficient ${\boldsymbol{\hat \alpha}}_j$, the set of similar patches $\hat{\mathcal{S}}_j$ and the weights $\hat{w}_i^j$ derived from the above approximations into the right-hand side of Eq.(\ref{eq:ObjFuncXDerivativeEdge}), such that: 
\begin{equation}
    \begin{array}{l}
        \big[({\mathbf H}_{k+1}^{\rm T}{\mathbf H}_{k+1}+\lambda_g){\mathbf G} + (\lambda_c+\lambda_s)\frac{N}{|{\boldsymbol M}|}\sum\limits_{j\in {\boldsymbol M}}{\mathbf Q}_j^{\rm T} {\mathbf Q}_j\big]{\boldsymbol{\hat X}}_{k+1} =
        \\ {\mathbf H}_{k+1}^{\rm T}{\mathbf G}\boldsymbol{Y} + \lambda_c\frac{N}{|{\boldsymbol M}|}\sum\limits_{j\in {\boldsymbol M}}{\mathbf Q}_j^{\rm T}{\mathbf D}{\boldsymbol{ \hat \alpha}}_j + \lambda_s\frac{N}{|{\boldsymbol M}|}\sum\limits_{j\in {\boldsymbol M}}{\mathbf Q}_j^{\rm T}\sum\limits_{i\in \hat{\mathcal {S}}_j}\hat{w}_i^j {\mathbf R}_i{\boldsymbol{\hat X}}^{a}_{k}
    \end{array}.
    \label{eq:ObjFuncXDerivativeApproxEdge}
\end{equation}
Since it is a linear equation with respect to ${\boldsymbol{\hat X}}_{k+1}$, Eq.(\ref{eq:ObjFuncXDerivativeApproxEdge}) can be solved by direct matrix inversion or the conjugate gradient method. In our method, ${\boldsymbol{\hat X}}_{k+1}$ are updated by solving it using the bi-conjugate gradient (BICG) method.

{\textbf {4. Repeat steps 1-3 until convergence or for a fixed number of iterations.}}

\subsection{Implementation}
\label{ssec:model}

To speed up the convergence and handle of large blurs, following most existing methods, we estimate the blur kernel in a coarse-to-fine framework. We apply our alternating iterative minimization procedure described in Section \ref{ssec:optimization} to each of the levels of the image pyramid constructed from the blurred image $\boldsymbol{y}$. The blur kernel refinement starts from the coarsest level and works down to the finest level with the original image resolution. At the coarsest level, the latent image estimate is initialized with the observed blurry image. The intermediate latent image estimated at each coarser level is interpolated and then propagated to the next finer level as an initial estimate of the latent image to refine the blur kernel estimate in higher resolutions. 

Different from \cite{YuChang}, in which the dictionary is trained from patches randomly sampled from the down-scaled blurry image, in this paper, the dictionary is trained from edge patches sampled directly from the intermediate latent image estimated at the coarser scale, and iteratively updated once for each image scale during the solution. We do not pay attention to the sparsity of the entire image over the learned dictionary, but only the sparsity of edge patches in the image, for our sparsity regularizer prefers the sharp image to the blurred one only for edge patches.

Blind deconvolution in general involves two stages. The motion blur kernel $\boldsymbol{h}$ is firstly estimated by alternately updating the motion blur kernel $\boldsymbol{h}$ and the latent image $\boldsymbol{x}$. The intermediate latent images estimated during the iterations have no direct influence on the final deblurring result, and only affect this result indirectly by contributing to the refinement of the blur kernel estimate $\boldsymbol{\hat h}$. Then, the final deblurring result $\boldsymbol{\hat x}$ is recovered from the given blurry image $\boldsymbol{y}$ with the estimated blur kernel $\boldsymbol{\hat h}$ for the finest level by performing a variaty of non-blind deconvolution methods, such as fast TV-$\ell_1$ deconvolution \cite{XuJia}, sparse deconvolution \cite{LevinWeiss2009} and EPLL \cite{ZoranWeiss} \emph{etc.}.

We estimate the blur kernel $\boldsymbol{h}$ by the implementation of the pseudo-code outlined in Algorithm \ref{alg:ObjFuncSolve}. We construct an image pyramid with $ L $ levels from the given blurry image $\boldsymbol{y}$. The number of pyramid levels is chosen such that, at the coarsest level, the size of the blur is smaller than that of the patch used in the blur kernel estimation stage. Let us use the notation ${\boldsymbol{\hat x}}_{k}^{l}$ for the intermediate latent image estimate, where the superscript $ l $ indicates the $ l $th level in the image pyramid, while the subscript $ k $ indicates the $ k $th iteration at each scale level. The iterative procedure starts from the coarsest level $ l=1 $ of the image pyramid initialized with ${\boldsymbol{\hat x}}_{0}^{1} = \boldsymbol{y}$. At each scale level $ l\in \{1,\cdots, L\} $, we take the iterative procedure that alternately optimizes the motion blur kernel $\boldsymbol{h}$ and the latent image $\boldsymbol{x}$ as detailed in Section \ref{ssec:optimization}, which is implemented repeatedly until the convergence or for a fixed number of iterations. Then the outcome of updating the latent image at the $ l $th level is upsampled by interpolation and then used as an initial estimate of the latent image for the next finer level $ {l+1} $ to progressively refine the motion blur kernel estimate $\boldsymbol{\hat h}$, which is repeated to achieve the final refinement of the blur kernel estimate $ {\boldsymbol{\hat h}} $ for the finest level. 

\begin{algorithm}[htbp]
    \caption{Edge-Based Blur Kernel Estimation Using Sparse Representation and Self-Similarity}
    \label{alg:ObjFuncSolve}
%    \begin{algorithmic}[1]
        \KwIn {Blurry image $\boldsymbol{y}$}
        \KwOut {Blur kernel estimate $\hat{\boldsymbol{h}}$}
        Set down-scaling factor $a$, regularization weights $\lambda_g$, $\lambda_c$, $\lambda_s$, $\lambda_h$, size of patch $n$, size of dictionary $t$, sparsity constraint parameter $T$, number of similar patches $p$, convergence tolerance $\epsilon$ and maximum allowed number of iterations ${\rm maxIters}$\;
        Build an image pyramid with $L$ levels\;
        Initialize $\hat{\boldsymbol{x}}_0^1 = \boldsymbol{y}$\;
        Train dictionary ${\mathbf D}$ using $ \hat{\boldsymbol{x}}_0^1 $\;
            \textbf{Outer loop:}
            \For (\tcp*[f]{\footnotesize {for each level of image pyramid}}) {$l=1$ to $l=L$}  
            {
            Initialize $k=0$, gradient threshold $ \tau $\;
                \textbf{Inner loop:} 
                \Repeat (\tcp*[f]{\footnotesize {for each iteration}}) { $k > {\rm maxIters}$ or $\Vert{\boldsymbol{\hat x}}_{k}^{l} - {\boldsymbol{\hat x}}_{k-1}^{l} \Vert_2^2\leqslant\epsilon $ }  
                {
                Predict the edge mask ${\boldsymbol {M}}$\;
                Compute blur kernel ${\boldsymbol{\hat h}}_{k+1}^{l}$ using Eq.(\ref{eq:ObjFuncSolveKEdge})\;
                Given ${\boldsymbol{\hat x}}_{k}^{l}$, update latent image ${\boldsymbol{\hat x}}_{k+1}^{l}$ by solving Eq. (\ref{eq:ObjFuncXDerivativeApproxEdge}) using BICG\;
                $\tau=\tau/1.1$; $k=k+1$
                }
            Update dictionary ${\mathbf D}$ using ${\boldsymbol{\hat x}}_{k}^{l}$\;
            Upscale image ${\boldsymbol{\hat x}}_{k}^{l}$ to initialize ${\boldsymbol{\hat x}}_{0}^{l+1}$ for the next finer level\;           
            }
        $ {\boldsymbol{\hat h}}={\boldsymbol{\hat h}}_{k}^{L} $;
        $ {\boldsymbol{\hat x}}={\boldsymbol{\hat x}}_{k}^{L} $.
%    \end{algorithmic}
\end{algorithm}

In the blur kernel estimation process, we use the gray-scale versions of the blurry image $\boldsymbol{y}$ and the intermediate latent image estimate $\boldsymbol{\hat x}$. Once the blur kernel estimate $\boldsymbol{\hat h}$ has been obtained with the original image scale, we perform the final non-blind deconvolution with $\boldsymbol{\hat h}$ on each color channel of $\boldsymbol{y}$ to obtain the deblurring result.

Finally, our method need perform deconvolution in the Fourier domain. To avoid ringing artifacts at the image boundaries, we process the image near the boundaries using the simple \emph{edgetaper} command in Matlab.
\section{EXPERIMENTS}
\label{sec:results}
Several experiments are conducted to demonstrate the performance of our method. We first test our method on the widely used datasets introduced in \cite{LevinWeiss2009} and \cite{SunCho}, and make qualitative and quantitative comparisons with the state-of-the-art blind deblurring methods. Then we show visual comparisons on real blurry photographs with unknown blurs. 
The relevant parameters of our method are set as follows: the dictionary $\mathbf{D}$ is of size $t=100$, and the sparsity constraint parameter $T=4$, designed to handle image patches of size $n=5\times5$, the number of iterations  is fixed as $14$ for the inner loop, and the regularization weights are empirically set to $\lambda_c=0.04/n$, $\lambda_s=0.04/n$, $\lambda_g=0.003$ and $\lambda_h=0.0003N$.
As the down-scaling factor increases, image patches become sharper, but there exist less similar patches at the down-sampled scale. Following the setting of \cite{MichaeliIrani}, the image pyramid is constructed with scale-gaps of $a = 4/3$ using down-scaling with a sinc function. Additional speed up is obtained by using the fast approximate nearest neighbor (NN) search of \cite{OlonetskyAvidan} in the blur kernel estimation stage, working with a single NN for every patch.

An additional important parameter is the size of the blur kernel. Small blurs are hard to solve if it is initialized with a very large kernel. Conversely, large blurs will be truncated if too small a kernel is used \cite{FergusSingh}. Following the setting of \cite{SunCho}, we do not assume that the size of the kernel is known and initialize that the size of the kernel is $51 \times 51$ in most cases except for some extremely difficult cases. Experiment results on both simulated and real blurry images show the size of the blur kernel is generally not larger than $51 \times 51$ for most blurry images. Even though the input blurry image has a small blur kernel, our method is still able to obtain a good deblurring result, less sensitive to the initial setting of the kernel size. 
\subsection{Quantitative Evaluation with Reference to Ground Truth}
\label{ssec:synthetic}
We test our method on two publicly available datasets. One dataset, which is provided by Levin et al. \cite{LevinWeiss2009}, contains $32$ images of size $255 \times 255$ blurred by real camera shake. The blurred images with spatially invariant blur and $8$ different ground-truth kernels were captured simultaneously by locking the Z-axis rotation handle but loosening the X and Y handles of the tripod. The kernels range in size from $13 \times 13$ to $27 \times 27$. 
%\begin{figure*}[htbp]
%    \centering
%    \begin{minipage}{0.24\textwidth}
%        \includegraphics[width=\textwidth]{levin_1}
%        \vskip1mm
%        \includegraphics[width=0.49\textwidth]{levin_k1}
%        \includegraphics[width=0.49\textwidth]{levin_k2}
%    \end{minipage}
%    \begin{minipage}{0.24\textwidth}
%        \includegraphics[width=\textwidth]{levin_2}
%        \vskip1mm
%        \includegraphics[width=0.49\textwidth]{levin_k3}
%        \includegraphics[width=0.49\textwidth]{levin_k4}
%    \end{minipage}
%    \begin{minipage}{0.24\textwidth}
%        \includegraphics[width=\textwidth]{levin_3}
%        \vskip1mm
%        \includegraphics[width=0.49\textwidth]{levin_k5}
%        \includegraphics[width=0.49\textwidth]{levin_k6}
%    \end{minipage}
%    \begin{minipage}{0.24\textwidth}
%        \includegraphics[width=\textwidth]{levin_4}
%        \vskip1mm
%        \includegraphics[width=0.49\textwidth]{levin_k7}
%        \includegraphics[width=0.49\textwidth]{levin_k8}
%    \end{minipage}
%    \caption{Ground truth images and kernels provided by Levin et al. \cite{LevinWeiss2009}}
%    \label{fig:levin_database}
%\end{figure*}
The other dataset provided by Sun et al. \cite{SunCho} comprises $640$ natural images of diverse scenes, which were obtained by synthetically blurring $80$ high-resolution images with the $8$ blur kernels from \cite{LevinWeiss2009} and adding $1\%$ white Gaussian noise to the blurred images. We present qualitative and quantitative comparisons with the state-of-the-art blind deblurring methods \cite{FergusSingh,ChoLee,XuJia,LevinWeiss2011,PerroneFavaro,MichaeliIrani,SunCho,ChoParis,KrishnanTay,PerroneDiethelm}.

We measure the quality of the blur kernel estimate $\hat{\boldsymbol{h}}$ using the error ratio measure ${\rm ER}$ \cite{MichaeliIrani}:
\begin{equation}
     {\rm ER} = \frac{\Vert\boldsymbol{x}-\hat{\boldsymbol{x}}_{\hat{\boldsymbol{h}}}\Vert_2^2}{\Vert\boldsymbol{x}-\hat{\boldsymbol{x}}_{\boldsymbol{h}}\Vert_2^2},
    \label{eq:ErrorRation}
\end{equation}
where $\hat{\boldsymbol{x}}_{\hat{\boldsymbol{h}}}$ corresponds to the deblurring result with the recovered kernel $\hat{\boldsymbol{h}}$, and $\hat{\boldsymbol{x}}_{\boldsymbol{h}}$ corresponds to the deblurring result with the ground-truth kernel $\boldsymbol{h}$. The smaller ${\rm ER}$ corresponds to the better quality. In principle, if ${\rm ER} = 1$, the recovered kernel yields a deblurring result as good as the ground-truth kernel.

\begin{figure}[htbp]
    \centering
    \includegraphics[width=0.99\columnwidth]{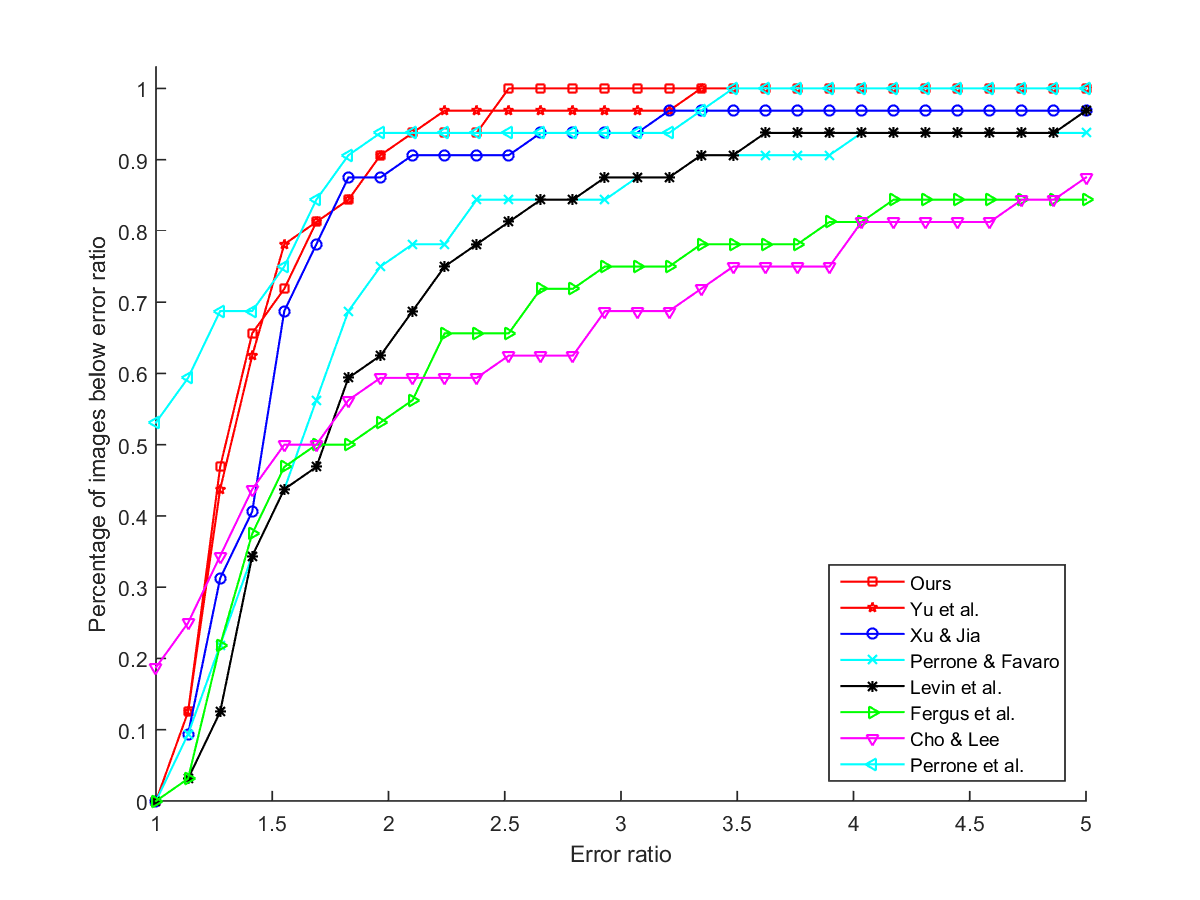}
    %\vskip1mm
    \caption{
    Cumulative distributions of error ratios with different methods on the dataset of \cite{LevinWeiss2009}
    }
    \label{fig:ERonLevin}
\end{figure}

On the dataset provided by Levin et al. \cite{LevinWeiss2009}, we compare our error ratios with those of Fergus et al. \cite{FergusSingh}, Cho and Lee \cite{ChoLee}, Xu and Jia \cite{XuJia}, Perrone and Favaro \cite{PerroneFavaro}, Levin et al. \cite{LevinWeiss2011}, Perrone et al. \cite{PerroneDiethelm} and our previous method \cite{YuChang}. Fig.\ref{fig:ERonLevin} shows the cumulative distribution of the error ratio of our method compared with the other methods over the dataset of \cite{LevinWeiss2009}. Levin et al. \cite{LevinWeiss2011} use sparse deconvolution \cite{LevinWeiss2009} to generate the final results, and observe that deconvolution results are usually visually plausible when their error ratios are below 3. Therefore, we standardize the final non-blind deconvolution by using sparse deconvolution \cite{LevinWeiss2009} to obtain the results, for fair comparison. 
Table \ref{tab:ER_result_levin} lists the success rate and the average error ratio over $32$ images for each method. The success rate is the percentage of images which achieve good deblurring results, that is, the percentage of images that have an error ratio below a certain threshold. On this dataset, the success rate is the percentage of the results under the error ratio of $3$. Table \ref{tab:ER_result_levin} shows our method takes the lead with a success rate of $ 100\% $, a higher success rate than our previous method without considering salient edges \cite{YuChang}. Levin et al. \cite{LevinWeiss2011}, Perrone and Favaro \cite{PerroneFavaro} and Perrone et al. \cite{PerroneDiethelm} initialize the size of the blur kernel with ground truth, while the size of the blur kernel is unknown for real scenes. Even so, our method still achieves a much higher success rate than the other methods over the dataset of \cite{LevinWeiss2009}.
%\clearpage
%
\begin{table}[htbp]
	\vbox{\centering
    \caption{Quantitative comparison of different methods over the dataset of \cite{LevinWeiss2009}}
    \label{tab:ER_result_levin}
    \vskip1mm
    \renewcommand{\baselinestretch}{1.35}
	{\footnotesize\centerline{\tabcolsep=15pt\begin{tabular}{ccc}
		\toprule[1pt]
			                                 &Success rate\%   & Mean error ratio\\
		\midrule[0.5pt]
		Ours                                      & 100        & 1.4433   \\
		Yu et al. \cite{YuChang}                  & 96.88      & 1.4653    \\
		Perrone et al. \cite{PerroneDiethelm}      & 93.75      & 1.2024   \\
		Xu \& Jia \cite{XuJia}                     & 93.75      & 2.1365    \\
        Perrone \& Favaro \cite{PerroneFavaro}     & 87.50      & 2.0263      \\ 
		Levin et al. \cite{LevinWeiss2011}         & 87.50      & 2.0583    \\
        Fergus et al. \cite{FergusSingh}           & 75.00      & 13.5268   \\
		Cho \& Lee \cite{ChoLee}                   & 68.75      & 2.6688    \\
		\bottomrule[1pt]
	\end{tabular}}}}
\end{table}

\begin{figure}[htbp]
    \centering
    \includegraphics[width=0.99\columnwidth]{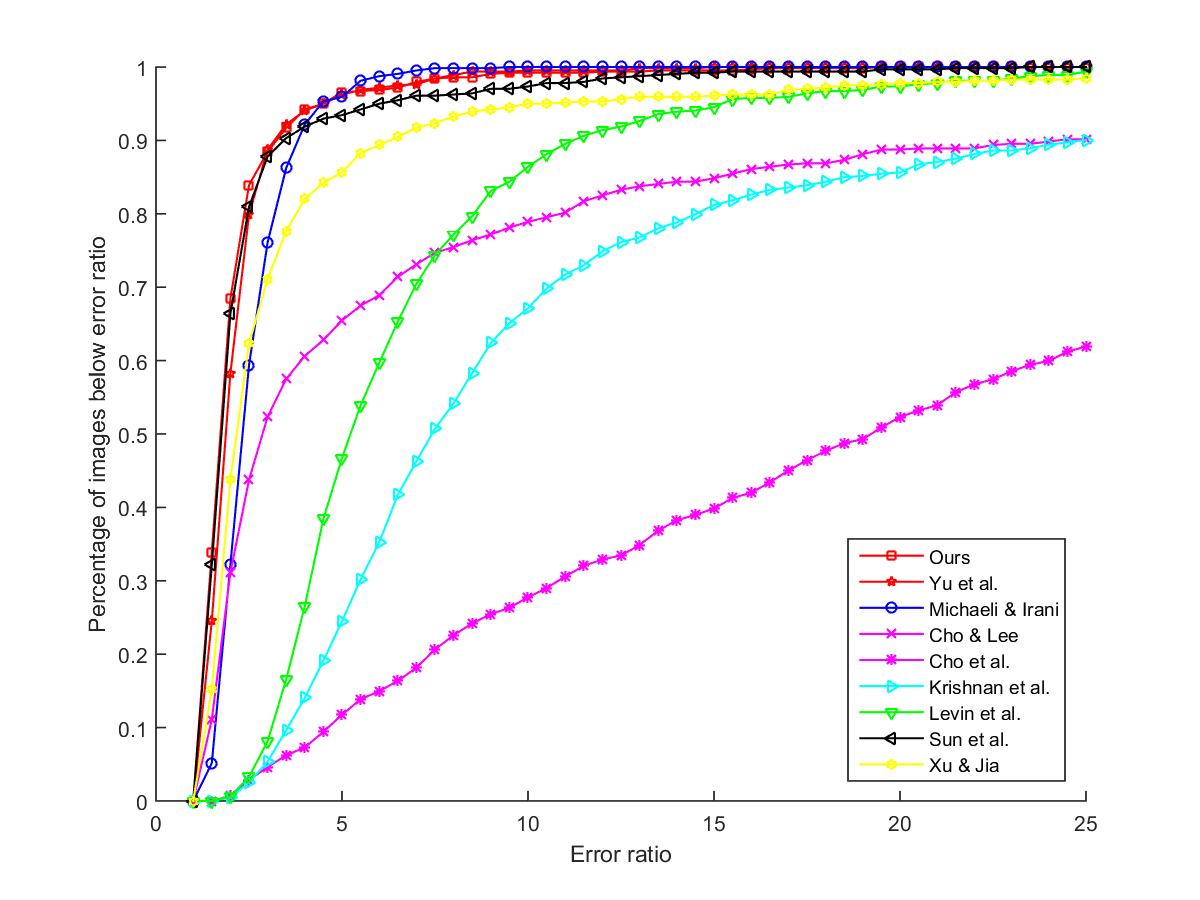}
    %\vskip2mm
    \caption{
    Cumulative distributions of error ratios with different methods on the dataset of \cite{SunCho}
    }
    \label{fig:ERonSun}
\end{figure}

On this dataset provided by Sun et al. \cite{SunCho}, we compare our error ratios with those of Cho and Lee \cite{ChoLee}, Xu and Jia \cite{XuJia}, Levin et al. \cite{LevinWeiss2011}, Sun et al. \cite{SunCho}, Michaeli and Irani \cite{MichaeliIrani}, Cho et al. \cite{ChoParis}, Krishnan et al. \cite{KrishnanTay} and our previous method \cite{YuChang}. Fig.\ref{fig:ERonSun} shows the cumulative distribution of error ratios over the entire dataset for each method. We apply the blur kernel estimated by each method to perform deblurring with the non-blind deblurring method of \cite{ZoranWeiss} to recover latent images. It is empirically observed by Michaeli and Irani \cite{MichaeliIrani} that the deblurring results are still visually acceptable for error ratios ${\rm ER} \leqslant 5$, when using the non-blind deconvolution of \cite{ZoranWeiss}. Table \ref{tab:ER_result_Sun} lists the success rate (\emph{i.e.}, an error ratio below $5$) and the average error ratio over $640$ images with different methods. Table \ref{tab:ER_result_Sun} shows our method achieves the highest success rate and the lowest average error ratio followed by Michaeli and Irani \cite{MichaeliIrani} and Sun et al. \cite{SunCho}. Moreover, these two methods by Michaeli and Irani \cite{MichaeliIrani} and Sun et al. \cite{SunCho} take 9213 and 4899 seconds on average to process an image of size $1024 \times 800$ from this dataset respectively, and our method take 1823 seconds, much faster than their methods. 
%\vskip2mm
%\vskip1mm
\begin{table}[htbp]
\vbox{\centering
\caption{Quantitative comparison of different methods over the dataset of \cite{SunCho}}
\label{tab:ER_result_Sun}
%\centering{\caption{Quantitative comparison of different methods over the dataset \cite{SunCho}}\label{tab:ER}}
\vskip1mm
\renewcommand{\baselinestretch}{1.35}
{\footnotesize\centerline{\tabcolsep=15pt\begin{tabular}{ccc}
\toprule[1pt]
                                      &Success rate\%   &Mean error ratio \\
\midrule[0.5pt]
Ours                                     & 96.56         & 2.1134    \\
Yu et al. \cite{YuChang}                 & 96.25         & 2.2047    \\
Michaeli \& Irani \cite{MichaeliIrani}    & 95.94         & 2.5662    \\
Sun et al. \cite{SunCho}                  & 93.44         & 2.3764    \\
Xu \& Jia \cite{XuJia}                    & 85.63         & 3.6293    \\
Levin et al. \cite{LevinWeiss2011}        & 46.72         & 6.5577    \\
Cho \& Lee \cite{ChoLee}                  & 65.47         & 8.6901    \\
Krishnan et al. \cite{KrishnanTay}        & 24.49         & 11.5212   \\
Cho et al. \cite{ChoParis}                & 11.74         & 24.7020   \\
\bottomrule[1pt]
\end{tabular}}}
\vskip0mm}
\end{table}

\begin{figure}[htbp]
    \centering
    \subfigure[{Blurry image}]
    {\begin{minipage}{0.1925\textwidth}
        \centering
        \includegraphics[width=0.99\textwidth,clip=true]{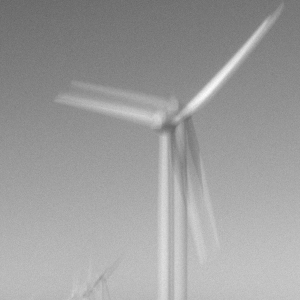}
        \vspace{1mm}
    \end{minipage}}
    \subfigure[{Ground truth}]
    {\begin{minipage}{0.1925\textwidth}
        \centering
        \includegraphics[width=0.99\textwidth,clip=true]{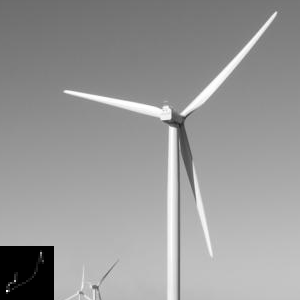}
        \vspace{1mm}
    \end{minipage}}
    \subfigure[{Cho and Lee \cite{ChoLee}}]
    {\begin{minipage}{0.1925\textwidth}
        \centering
        \includegraphics[width=0.99\textwidth,clip=true]{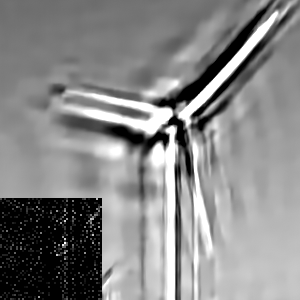}
        \vspace{1mm}
    \end{minipage}}
    \subfigure[{Xu \& Jia \cite{XuJia}}]
    {\begin{minipage}{0.1925\textwidth}
        \centering
        \includegraphics[width=0.99\textwidth,clip=true]{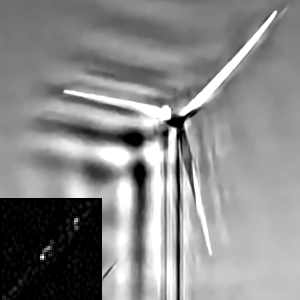}
        \vspace{1mm}
    \end{minipage}}
    \subfigure[{Krishnan et al. \cite{KrishnanTay}}]
    {\begin{minipage}{0.1925\textwidth}
        \centering
        \includegraphics[width=0.99\textwidth,clip=true]{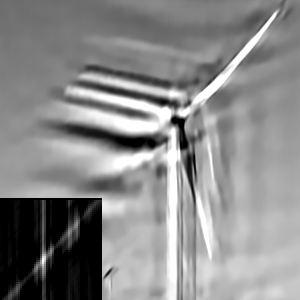}
        \vspace{1mm}
    \end{minipage}}
    \subfigure[{Cho et al. \cite{ChoParis}}]
    {\begin{minipage}{0.1925\textwidth}
        \centering
        \includegraphics[width=0.99\textwidth,clip=true]{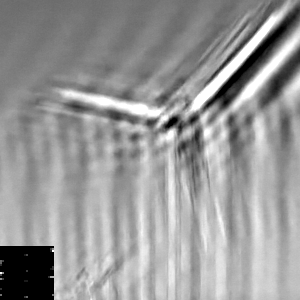}
        \vspace{1mm}
    \end{minipage}}
    \subfigure[{Levin et al. \cite{LevinWeiss2011}}]
    {\begin{minipage}{0.1925\textwidth}
        \centering
        \includegraphics[width=0.99\textwidth,clip=true]{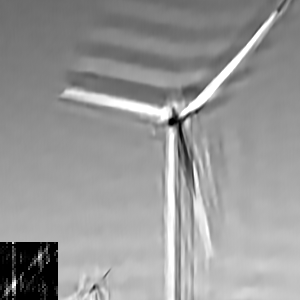}
        \vspace{1mm}
    \end{minipage}}
    \subfigure[{Sun et al. \cite{SunCho}}]
    {\begin{minipage}{0.1925\textwidth}
        \centering
        \includegraphics[width=0.99\textwidth,clip=true]{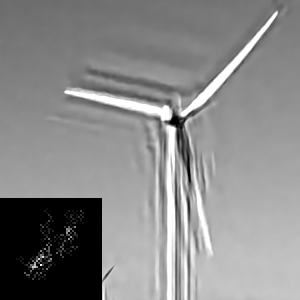}
        \vspace{1mm}
    \end{minipage}}
    \subfigure[{Michaeli \& Irani \cite{MichaeliIrani}}]
    {\begin{minipage}{0.1925\textwidth}
        \centering
        \includegraphics[width=0.99\textwidth,clip=true]{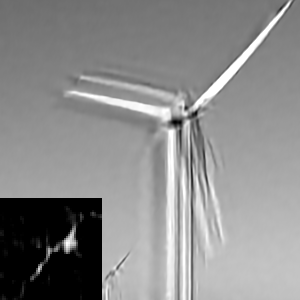}
        \vspace{1mm}
    \end{minipage}}    
%    \subfigure[{Yu et al. \cite{YuChang}}]
%    {\begin{minipage}{0.19\textwidth}
%        \centering
%        \includegraphics[width=0.99\textwidth,clip=true]{78_4_our_aas_update_D.png}
%        \vspace{1mm}
%    \end{minipage}}
    \subfigure[{Our method}]
    {\begin{minipage}{0.1925\textwidth}
        \centering
        \includegraphics[width=0.99\textwidth,clip=true]{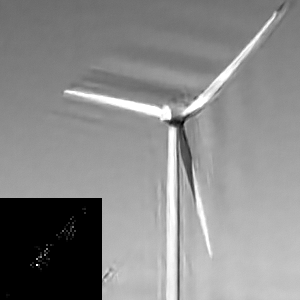}
        \vspace{1mm}
    \end{minipage}}
    %\vskip2mm
    \caption{Qualitative comparison of different methods on a cropped image from the synthetic dataset of \cite{SunCho}}
    \label{fig:simulres_78}
\end{figure}

\begin{figure}[htbp]
    \centering
    \subfigure[{Blurry image}]
    {\begin{minipage}{0.1925\textwidth}
        \centering
        \includegraphics[width=0.99\textwidth,clip=true]{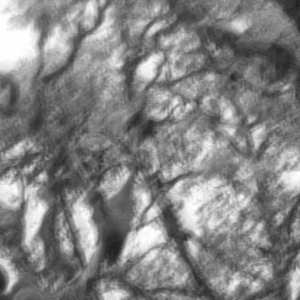}
        \vspace{1mm}
    \end{minipage}}
    \subfigure[{Ground truth}]
    {\begin{minipage}{0.1925\textwidth}
        \centering
        \includegraphics[width=0.99\textwidth,clip=true]{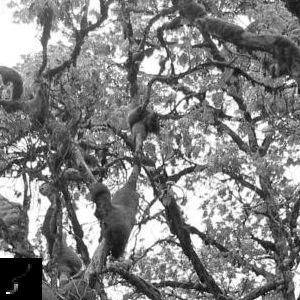}
        \vspace{1mm}
    \end{minipage}}
    \subfigure[{Cho and Lee \cite{ChoLee}}]
    {\begin{minipage}{0.1925\textwidth}
        \centering
        \includegraphics[width=0.99\textwidth,clip=true]{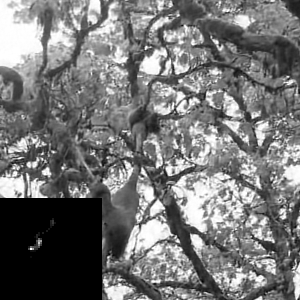}
        \vspace{1mm}
    \end{minipage}}
    \subfigure[{Xu \& Jia \cite{XuJia}}]
    {\begin{minipage}{0.1925\textwidth}
        \centering
        \includegraphics[width=0.99\textwidth,clip=true]{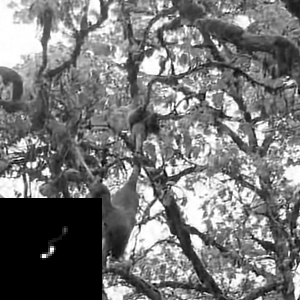}
        \vspace{1mm}
    \end{minipage}}
    \subfigure[{Krishnan et al. \cite{KrishnanTay}}]
    {\begin{minipage}{0.1925\textwidth}
        \centering
        \includegraphics[width=0.99\textwidth,clip=true]{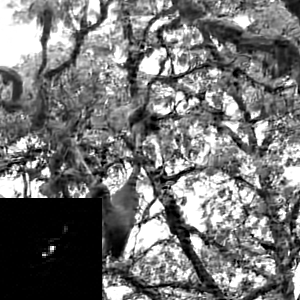}
        \vspace{1mm}
    \end{minipage}}\\
    \subfigure[{Cho et al. \cite{ChoParis}}]
    {\begin{minipage}{0.1925\textwidth}
        \centering
        \includegraphics[width=0.99\textwidth,clip=true]{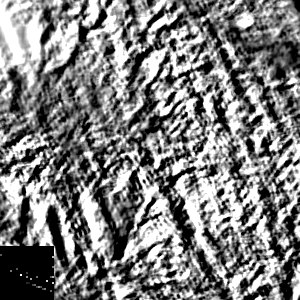}
        \vspace{1mm}
    \end{minipage}}
    \subfigure[{Levin et al. \cite{LevinWeiss2011}}]
    {\begin{minipage}{0.1925\textwidth}
        \centering
        \includegraphics[width=0.99\textwidth,clip=true]{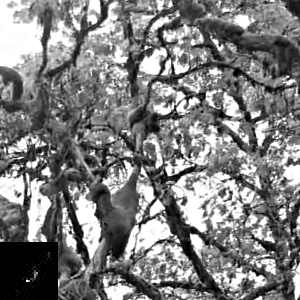}
        \vspace{1mm}
    \end{minipage}}
    \subfigure[{Sun et al. \cite{SunCho}}]
    {\begin{minipage}{0.1925\textwidth}
        \centering
        \includegraphics[width=0.99\textwidth,clip=true]{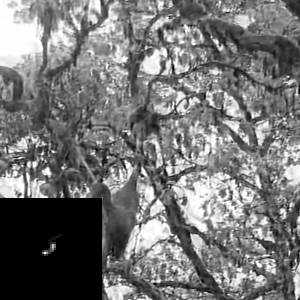}
        \vspace{1mm}
    \end{minipage}}
    \subfigure[{ Michaeli \& Irani \cite{MichaeliIrani}}]
    {\begin{minipage}{0.1925\textwidth}
        \centering
        \includegraphics[width=0.99\textwidth,clip=true]{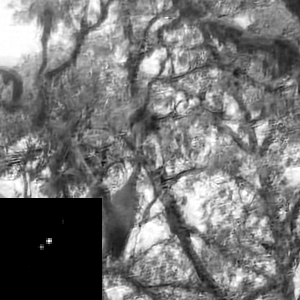}
        \vspace{1mm}
    \end{minipage}}    
%    \subfigure[{Yu et al. \cite{YuChang}}]
%    {\begin{minipage}{0.19\textwidth}
%        \centering
%        \includegraphics[width=0.99\textwidth,clip=true]{30_6_our_aas_update_D.png}
%        \vspace{1mm}
%    \end{minipage}}
    \subfigure[{Our method}]
    {\begin{minipage}{0.1925\textwidth}
        \centering
        \includegraphics[width=0.99\textwidth,clip=true]{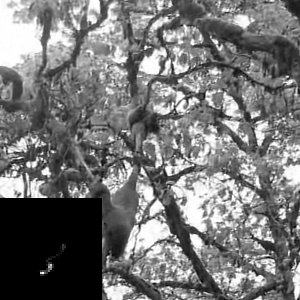}
        \vspace{1mm}
    \end{minipage}}
    %\vskip2mm
    \caption{Qualitative comparison of different methods on another cropped image from the synthetic dataset of \cite{SunCho}}
    \label{fig:simulres_30}
\end{figure}

Figs.\ref{fig:simulres_78} and \ref{fig:simulres_30} show qualitative comparisons of cropped results on two blurred images from the synthetic dataset of \cite{SunCho} by different methods. Compared with the other methods, our method usually obtains more accurate blur kernels, suffers from fewer ringing artifacts and restores more and sharper image details.

\subsection{Qualitative Comparison on Real Images}
\label{ssec:real}
\begin{figure}[htbp]
    \centering
    \subfigure[{Blurry image}]
    {\begin{minipage}{0.32\textwidth}
        \centering
        \includegraphics[width=\textwidth,clip=true]{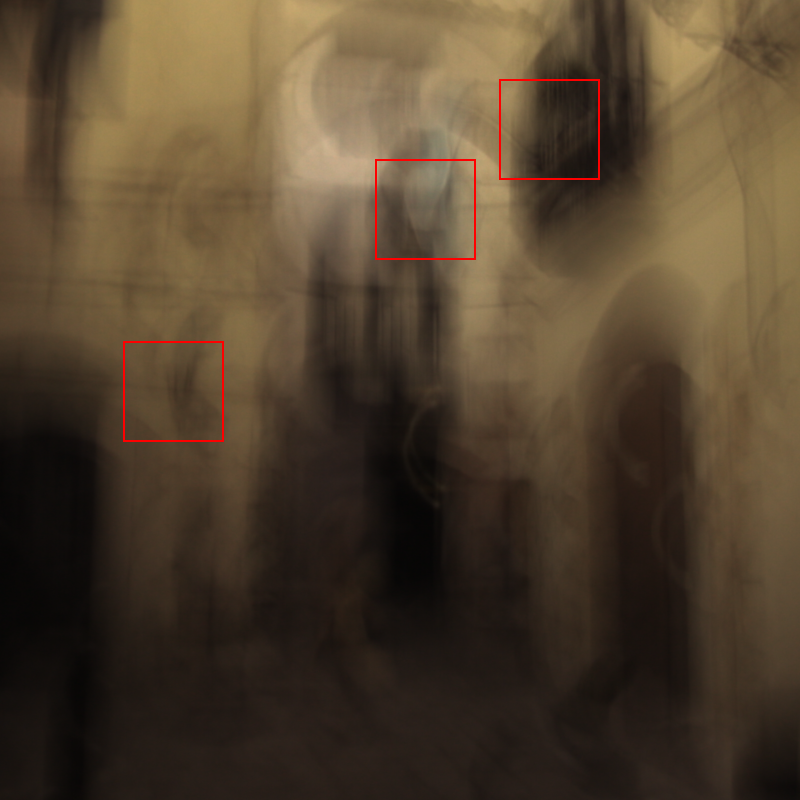}\\
        \includegraphics[width=\textwidth,clip=true]{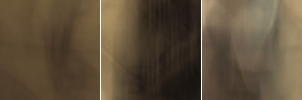}
        \vspace{1mm}
    \end{minipage}}
    \subfigure[{Xu \& Jia \cite{XuJia}}]
    {\begin{minipage}{0.32\textwidth}
        \centering
        \includegraphics[width=\textwidth,clip=true]{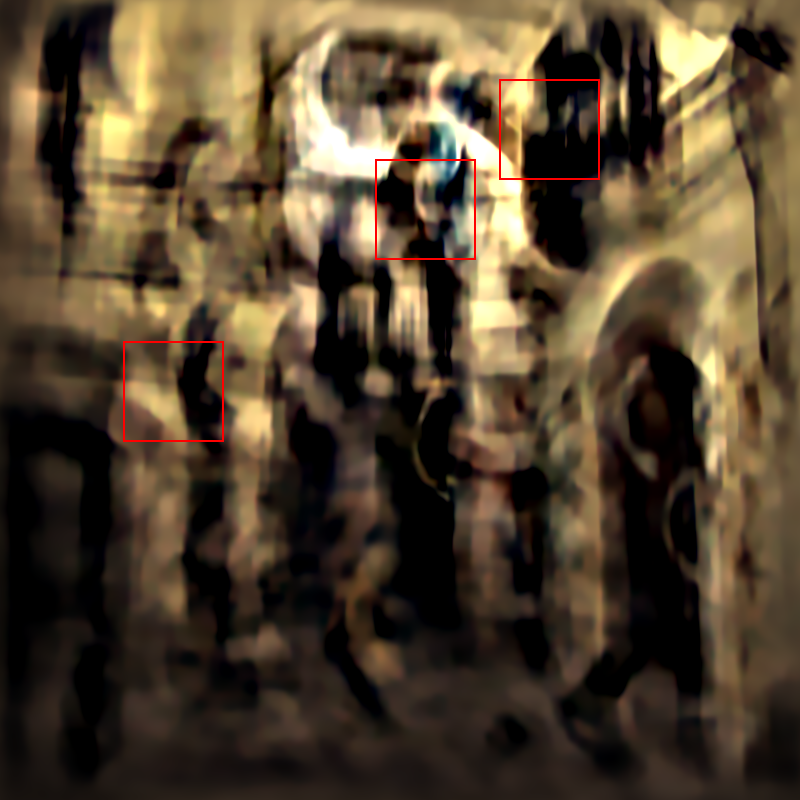}\\
        \includegraphics[width=\textwidth,clip=true]{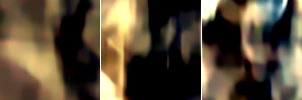}
        \vspace{1mm}
    \end{minipage}}
    \subfigure[{Krishnan et al. \cite{KrishnanTay}}]
    {\begin{minipage}{0.32\textwidth}
        \centering
        \includegraphics[width=\textwidth,clip=true]{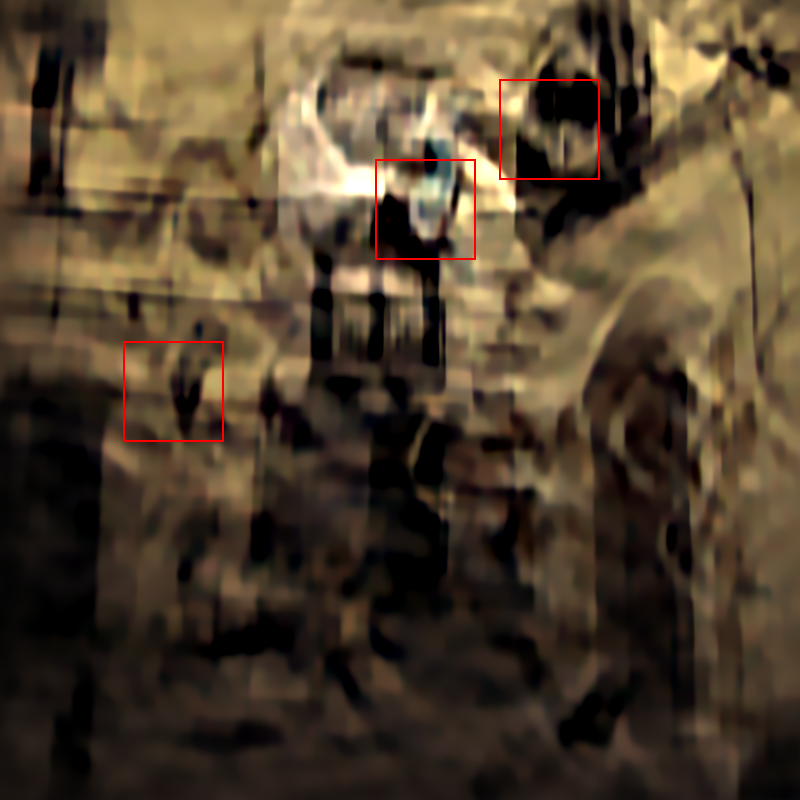}\\
        \includegraphics[width=\textwidth,clip=true]{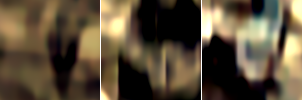}
        \vspace{1mm}
    \end{minipage}}
    \subfigure[{Sun et al. \cite{SunCho}}]
    {\begin{minipage}{0.32\textwidth}
        \centering
        \includegraphics[width=\textwidth,clip=true]{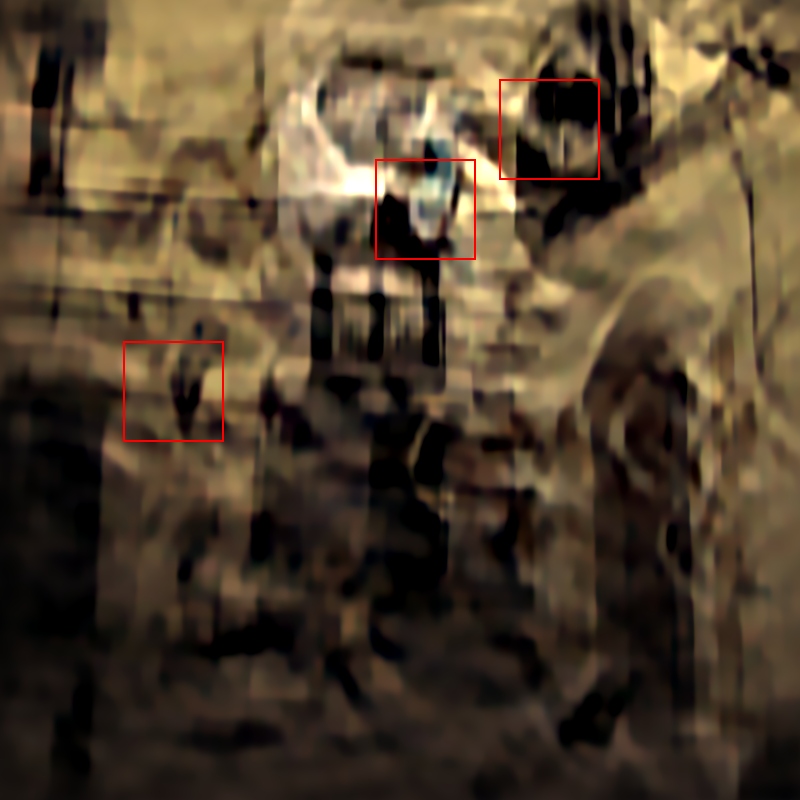}\\
        \includegraphics[width=\textwidth,clip=true]{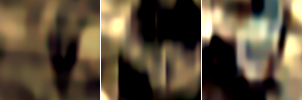}
        \vspace{1mm}
    \end{minipage}}
    \subfigure[{Perrone \& Favaro \cite{PerroneFavaro}}]
    {\begin{minipage}{0.32\textwidth}
        \centering
        \includegraphics[width=\textwidth,clip=true]{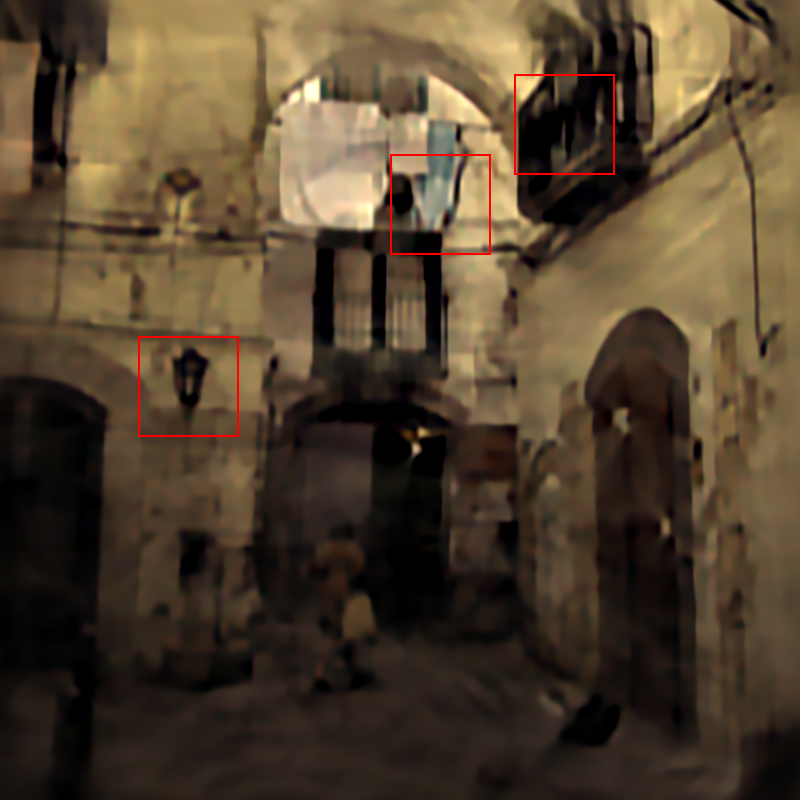}\\
        \includegraphics[width=\textwidth,clip=true]{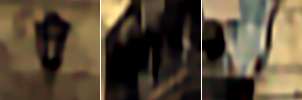}
        \vspace{1mm}
    \end{minipage}}
    \subfigure[{Perrone et al. \cite{PerroneDiethelm}}]
    {\begin{minipage}{0.32\textwidth}
        \centering
        \includegraphics[width=\textwidth,clip=true]{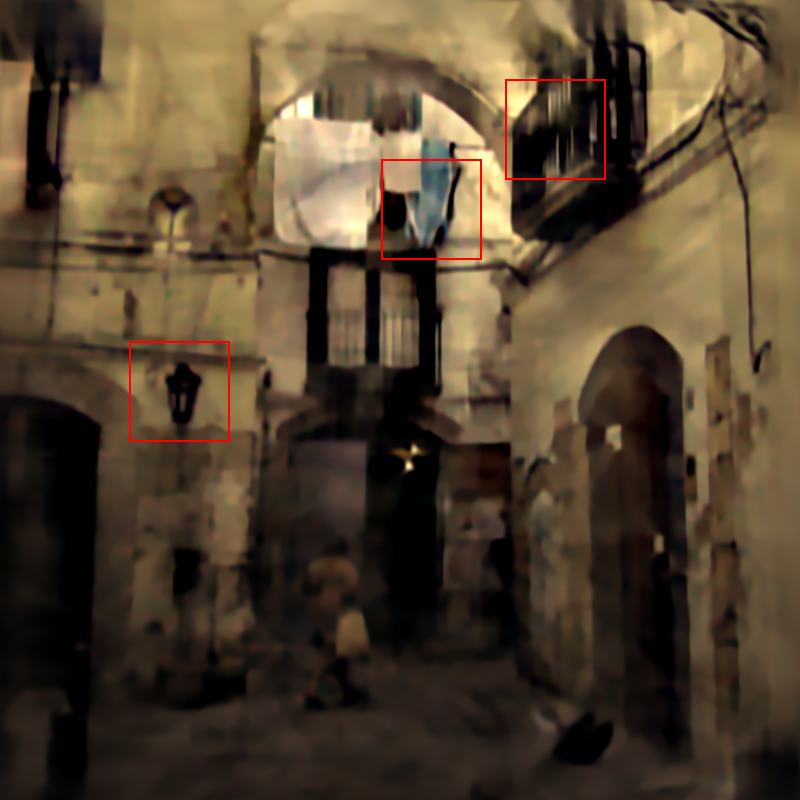}\\
        \includegraphics[width=\textwidth,clip=true]{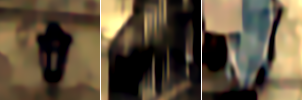}
        \vspace{1mm}
    \end{minipage}}
%    \subfigure[{Yu et al. \cite{YuChang}}]
%    {\begin{minipage}{0.49\textwidth}
%        \centering
%        \includegraphics[height=0.6\textwidth,clip=true]{Blurry3_8_Our_aas}\\
%        \includegraphics[height=0.6\textwidth,clip=true]{Blurry3_8_Our_aas_patch}
%        \vspace{1mm}
%    \end{minipage}}
    \subfigure[{Pan et al. \cite{PanSun}}]
    {\begin{minipage}{0.32\textwidth}
        \centering
        \includegraphics[width=\textwidth,clip=true]{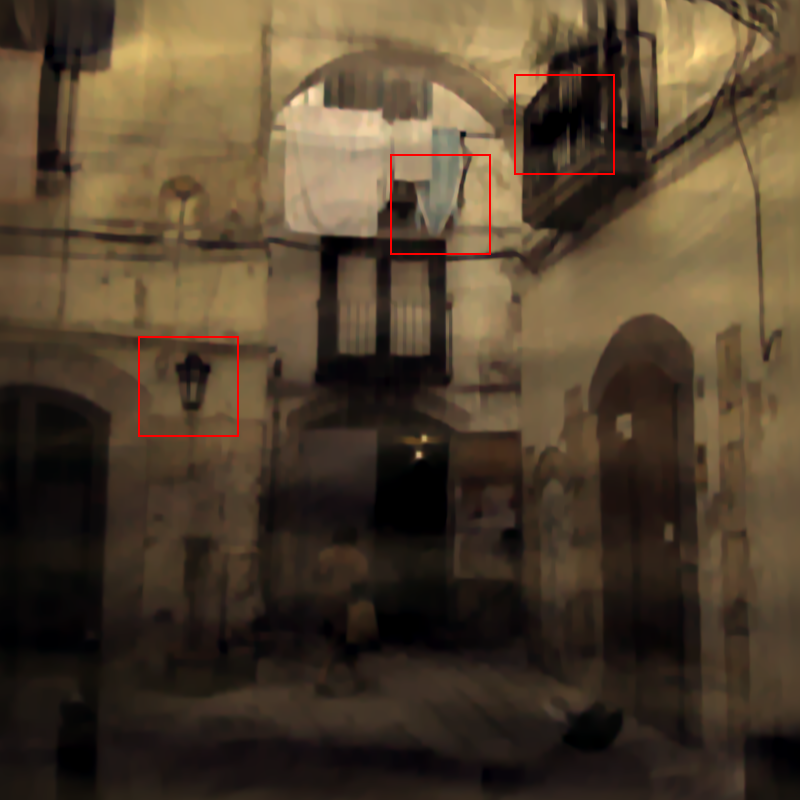}\\
        \includegraphics[width=\textwidth,clip=true]{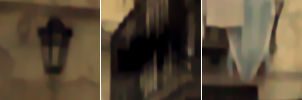}
        \vspace{1mm}
    \end{minipage}}
    \subfigure[{Yan et al. \cite{YanRen}}]
    {\begin{minipage}{0.32\textwidth}
        \centering
        \includegraphics[width=\textwidth,clip=true]{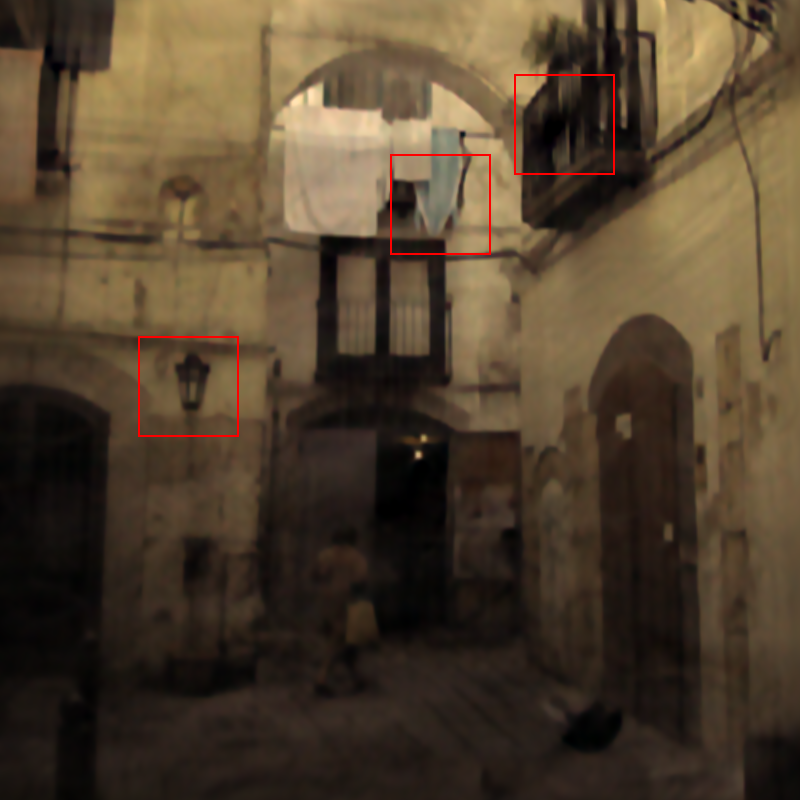}\\
        \includegraphics[width=\textwidth,clip=true]{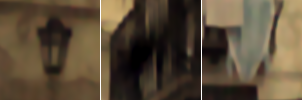}
        \vspace{1mm}
    \end{minipage}}
    \subfigure[{Our method}]
    {\begin{minipage}{0.32\textwidth}
        \centering
        \includegraphics[width=\textwidth,clip=true]{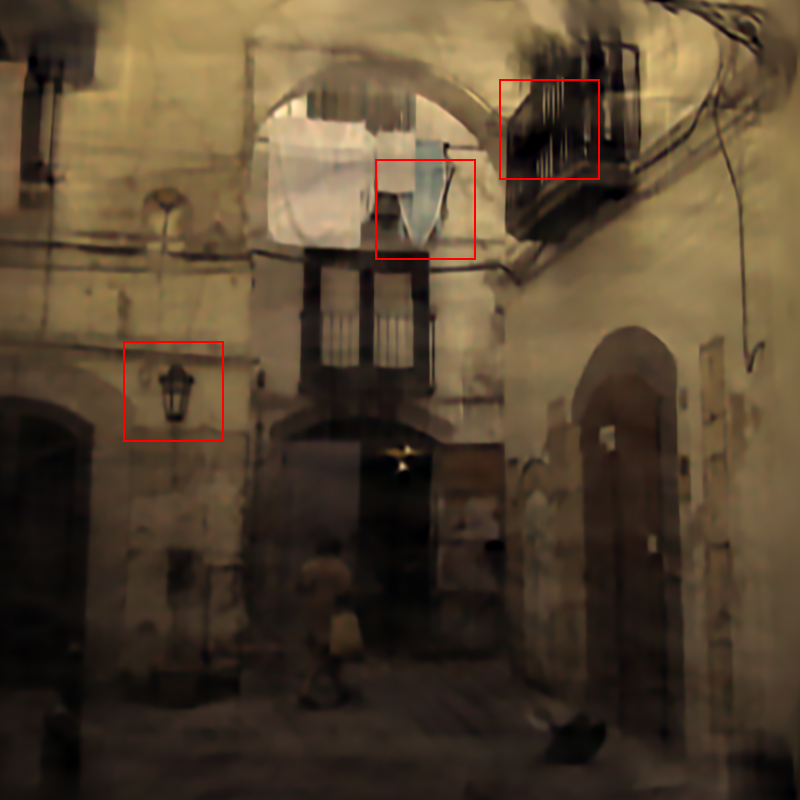}
        \includegraphics[width=\textwidth,clip=true]{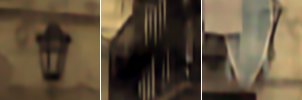}
        \vspace{1mm}
    \end{minipage}}
    \caption{Visual comparison between our method and some state-of-the-art methods on real blurry image with unknown large blur}
    \label{fig:res_Blurry3_8}
\end{figure}

\begin{figure}[htbp]
    \centering
    \subfigure[{Blurry image}]
    {\begin{minipage}{0.32\textwidth}
        \centering
        \includegraphics[width=\textwidth,clip=true]{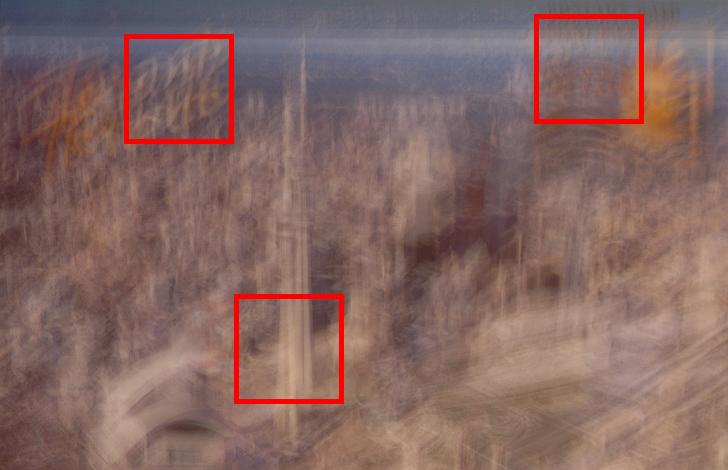}\\
        \includegraphics[width=\textwidth,clip=true]{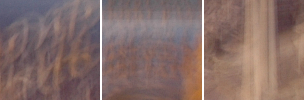}
        \vspace{1mm}
    \end{minipage}}
    \subfigure[{Xu \& Jia \cite{XuJia}}]
    {\begin{minipage}{0.32\textwidth}
        \centering
        \includegraphics[width=\textwidth,clip=true]{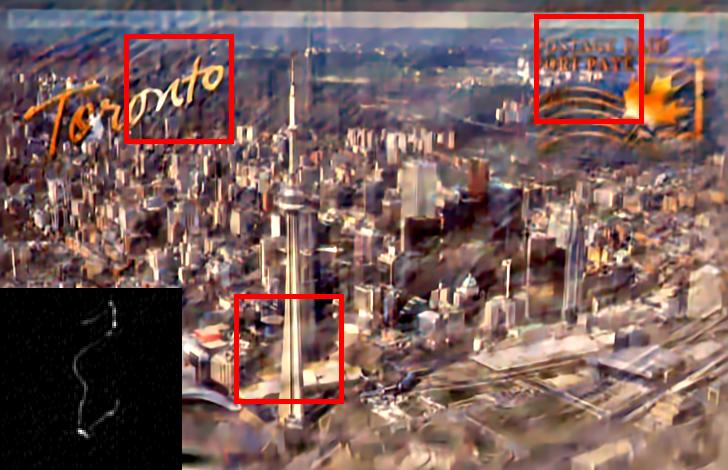}\\
        \includegraphics[width=\textwidth,clip=true]{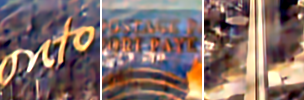}
        \vspace{1mm}
    \end{minipage}}
    \subfigure[{Krishnan et al. \cite{KrishnanTay}}]
    {\begin{minipage}{0.32\textwidth}
        \centering
        \includegraphics[width=\textwidth,clip=true]{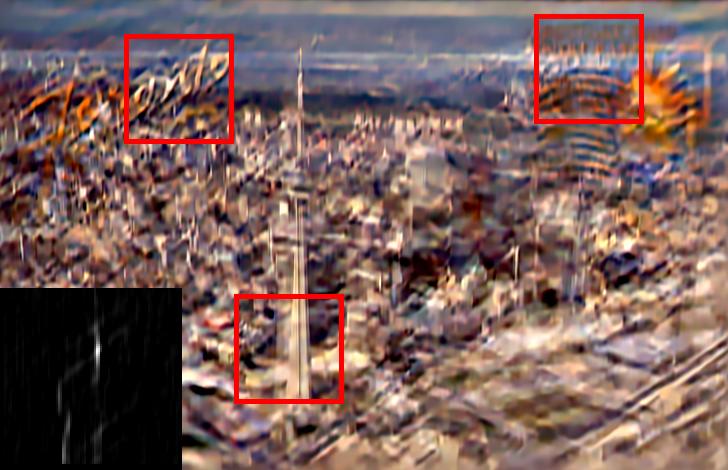}\\
        \includegraphics[width=\textwidth,clip=true]{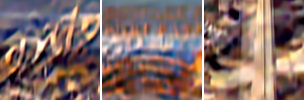}
        \vspace{1mm}
    \end{minipage}}
    \subfigure[{Sun et al. \cite{SunCho}}]
    {\begin{minipage}{0.32\textwidth}
        \centering
        \includegraphics[width=\textwidth,clip=true]{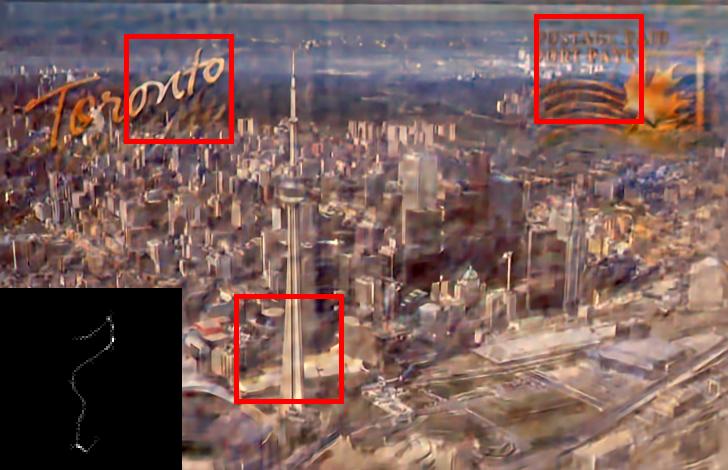}\\
        \includegraphics[width=\textwidth,clip=true]{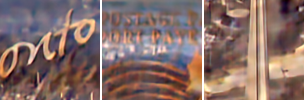}
        \vspace{1mm}
    \end{minipage}}
    \subfigure[{Michaeli \& Irani \cite{MichaeliIrani}}]
    {\begin{minipage}{0.32\textwidth}
        \centering
        \includegraphics[width=\textwidth,clip=true]{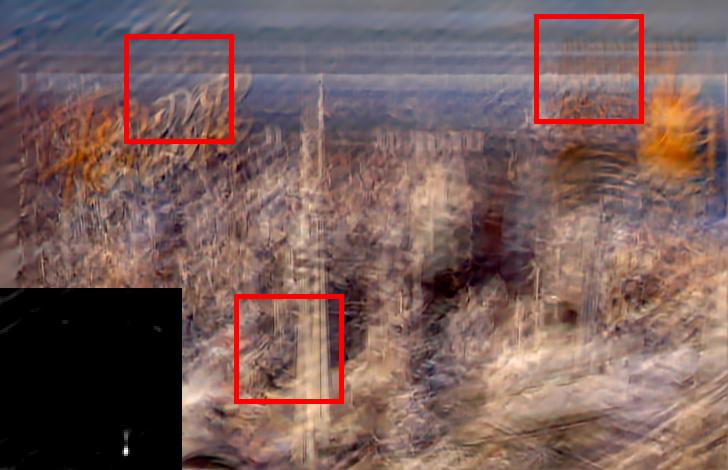}\\
        \includegraphics[width=\textwidth,clip=true]{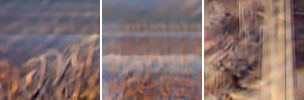}
        \vspace{1mm}
    \end{minipage}}
    \subfigure[{Perrone \& Favaro \cite{PerroneFavaro}}]
    {\begin{minipage}{0.32\textwidth}
        \centering
        \includegraphics[width=\textwidth,clip=true]{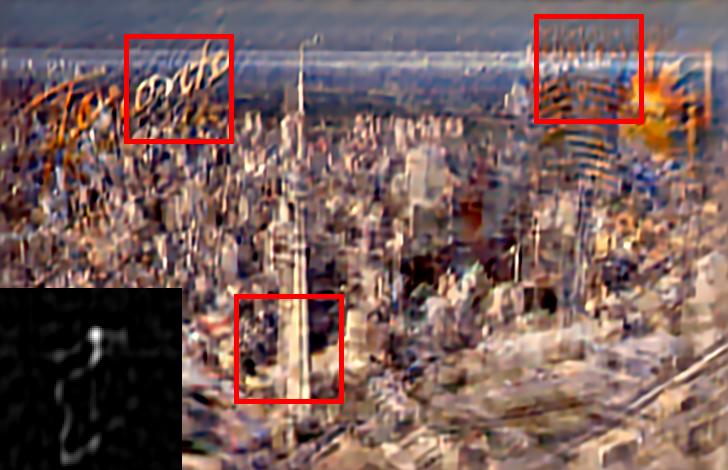}\\
        \includegraphics[width=\textwidth,clip=true]{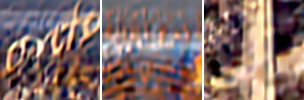}
        \vspace{1mm}
    \end{minipage}}
    \subfigure[{Perrone et al. \cite{PerroneDiethelm}}]
    {\begin{minipage}{0.32\textwidth}
        \centering
        \includegraphics[width=\textwidth,clip=true]{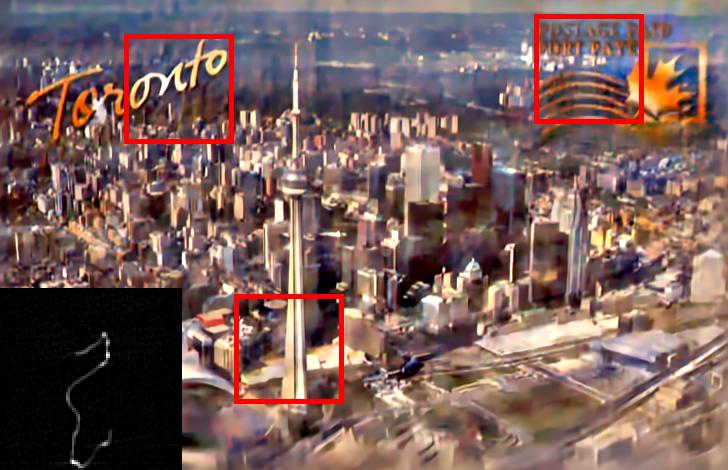}\\
        \includegraphics[width=\textwidth,clip=true]{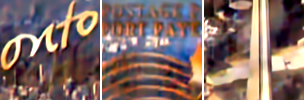}
        \vspace{1mm}
    \end{minipage}}
    \subfigure[{Yu et al. \cite{YuChang}}]
    {\begin{minipage}{0.32\textwidth}
        \centering
        \includegraphics[width=\textwidth,clip=true]{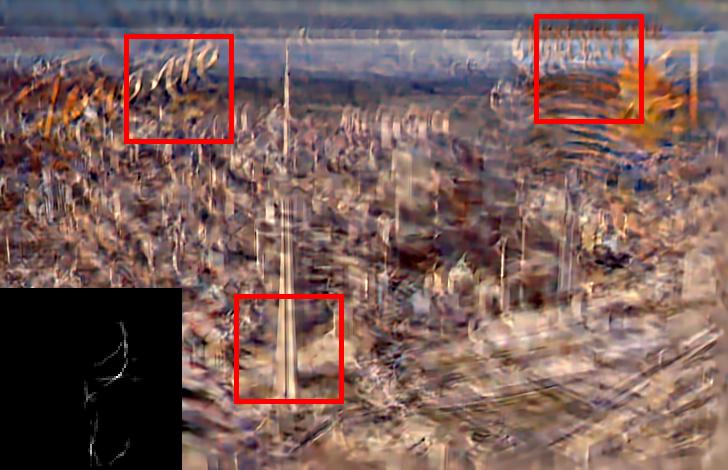}\\
        \includegraphics[width=\textwidth,clip=true]{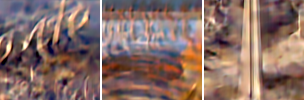}
        \vspace{1mm}
    \end{minipage}}
    \subfigure[{Our method}]
    {\begin{minipage}{0.32\textwidth}
        \centering
        \includegraphics[width=\textwidth,clip=true]{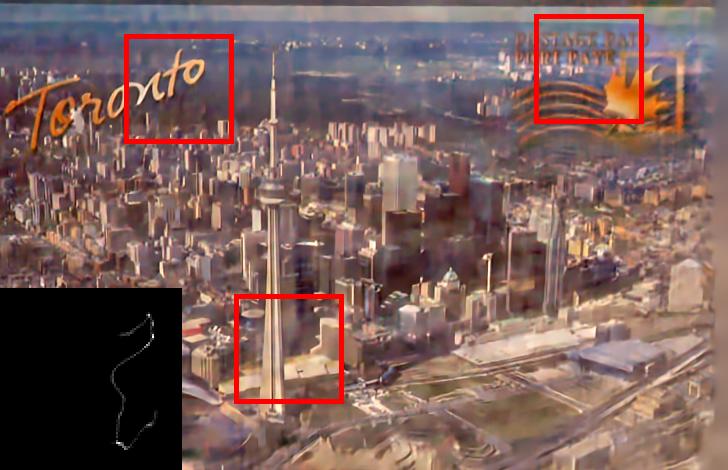}\\
        \includegraphics[width=\textwidth,clip=true]{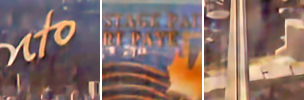}
        \vspace{1mm}
    \end{minipage}}
    \vskip2mm
    \caption{Visual comparison between our method and some state-of-the-art methods on another real blurry image with unknown large blur}
    \label{fig:res_postcard}
\end{figure}

We also experiment with real blurry images which are blurred with unknown kernels. In this part, we process blurry images with very large blurs to demonstrate the robustness of our method. We recover the latent image from the observed blurry image by performing the non-blind deconvolution method of \cite{ZoranWeiss} in the deblurring stage once the blur kernel has been estimated. Several methods are terminated early during the iteration due to lack of memory caused by too large the blur kernel.  
Fig.\ref{fig:res_Blurry3_8} shows a visual comparison example with the state-of-the-art blind deconvolution methods \cite{XuJia,KrishnanTay,SunCho,PerroneFavaro,PerroneDiethelm,PanSun,YanRen} on one blurred image from Kohler et al.'s dataset \cite{KohlerHirsch}, at the bottom of which are close-ups of different parts of these images. The results illustrate a noticeable contrast improvement that our method recovers sharper edges and more fine details with negligible artifacts, and achieves better visual quality, as it estimates more accurate blur kernels. 
We observe from Fig. \ref{fig:res_Blurry3_8} that the deblurred images by Perrone et al. \cite{PerroneFavaro,PerroneDiethelm} suffer from ringing artifacts, and some fine details such as the fence and the lantern are not properly recovered by Pan et al. \cite{PanSun} and Yan et al. \cite{YanRen}. Fig.\ref{fig:res_postcard} gives another visual comparison example with the state-of-the-art blind deconvolution methods \cite{XuJia,KrishnanTay,SunCho,MichaeliIrani,PerroneFavaro,PerroneDiethelm,YuChang}.
The size of the blur kernel can be automatically estimated in the pre-processing stage. In the above examples, the sizes of the blur kernels are empirically initialized to $151 \times 151$ and $91 \times 91$ respectively. Experimental results on real blurry photographs with unknown large blurs validate that our method is quite robust to deal with large blurs.

When the blur is close to or even wider than the edge, the structure of the sharp edge will significantly change after blur. For such a highly blurred image, insignificant edges do not always provide useful information and instead mistake the kernel estimation. Nevertheless, large-scale structures are confused slightly by the blur due to their salient edges and provide informative edges for blur kernel estimation. Accordingly, it is more reasonable to obtain an accurate estimate of the blur kernel relying on salient edges. For small blurs, most of the edges are wider than the blur kernel and all helpful for kernel estimation besides salient edges. In this case, the edge-based method proposed in this paper only has a slight improvement over our previous method without considering salient edges \cite{YuChang}. But for large blurs, since insignificant edges could disturb kernel estimation and only salient edges around large-scale structures help kernel estimation, the edge-based method can achieve much better deblurring results and successfully handle severely blurred images. 

%\vskip0mm

\section{Conclusion}
\label{sec:conclusion}
In this paper, we have presented an edge-based blur kernel estimation method for blind motion deblurring unifying sparse representation and self-similarity of edge patches as image priors to guide the recovery of the latent image. We construct the sparsity regularizer and the cross-scale non-local regularizer based on our patch priors, exploiting thoroughly prior knowledge from similar patches across different scales of the latent image, and incorporate these two regularizers into our blind deconvolution model. We find that our regularizers prefer the sharp image to the blurred one only around salient edges, and accordingly impose our regularizers on salient edge patches of the image for blur kernel estimation. We have extensively validated the performance of our method, and it is able to deblur images with excessively large blur kernels. 
% or


%\appendices

%\section{}

%\begin{acknowledgements}
%This work was supported by National Natural Science Foundation of China (61501008) and Beijing Municipal Natural Science Foundation (4172002).
%\end{acknowledgements}
\medskip
\noindent \textbf{Compliance with Ethical Standards}
\medskip
\\\noindent \textbf{Funding} This study was funded by National Natural Science Foundation of China (61501008) and Beijing Municipal Natural Science Foundation (4172002). 
\medskip
\\\noindent \textbf{Conflict of Interest} The authors declare that they have no conflicts of interest.

\bibliographystyle{elsarticle-num}
\bibliography{bibliography}   % name your BibTeX data base

\end{document}